\documentclass[10pt,twocolumn,letterpaper]{article}

\usepackage{authblk}
\usepackage[accsupp]{axessibility}  

\usepackage{wacv}              

\input{preamble}

\definecolor{wacvblue}{rgb}{0.21,0.49,0.74}
\usepackage[pagebackref,breaklinks,colorlinks,allcolors=wacvblue]{hyperref}

\usepackage[table]{xcolor}
\usepackage{multirow}
\usepackage{pifont}
\def\wacvPaperID{1614} 
\def\confName{WACV}
\def\confYear{2026}

\title{Power of Boundary and Reflection: Semantic Transparent Object Segmentation using Pyramid Vision Transformer with Transparent Cues}

\author{Tuan-Anh Vu\textsuperscript{1} \qquad Hai Nguyen-Truong\textsuperscript{1} \qquad Ziqiang Zheng\textsuperscript{1} \qquad 
Binh-Son Hua\textsuperscript{2} \\  Qing Guo\textsuperscript{3} \qquad Ivor W. Tsang\textsuperscript{4} \qquad Sai-Kit Yeung\textsuperscript{1}}

\affil{
\textsuperscript{1}Hong Kong University of Science and Technology \quad
\textsuperscript{2}Trinity College Dublin \\
\textsuperscript{3}Nankai University\quad 
\textsuperscript{4}Centre for Frontier AI Research (CFAR), A*STAR 
}

\begin{document}
\maketitle
\begin{abstract}
    Glass is a prevalent material among solid objects in everyday life, yet segmentation methods struggle to distinguish it from opaque materials due to its transparency and reflection. While it is known that human perception relies on boundary and reflective-object features to distinguish glass objects, the existing literature has not yet sufficiently captured both properties when handling transparent objects. Hence, we propose incorporating both of these powerful visual cues via the Boundary Feature Enhancement and Reflection Feature Enhancement modules in a mutually beneficial way. Our proposed framework, \textbf{TransCues}, is a pyramidal transformer encoder-decoder architecture to segment transparent objects. We empirically show that these two modules can be used together effectively, improving overall performance across various benchmark datasets, including glass object semantic segmentation, mirror object semantic segmentation, and generic segmentation datasets. Our method outperforms the state-of-the-art by a large margin, achieving \textbf{+4.2\%} mIoU on Trans10K-v2, \textbf{+5.6\%} mIoU on MSD, \textbf{+10.1\%} mIoU on RGBD-Mirror, \textbf{+13.1\%} mIoU on TROSD, and \textbf{+8.3\%} mIoU on Stanford2D3D, showing the effectiveness of our method against glass objects.
\end{abstract}

\begin{figure}[!t]
    \centering
    \includegraphics[width=0.9\linewidth]{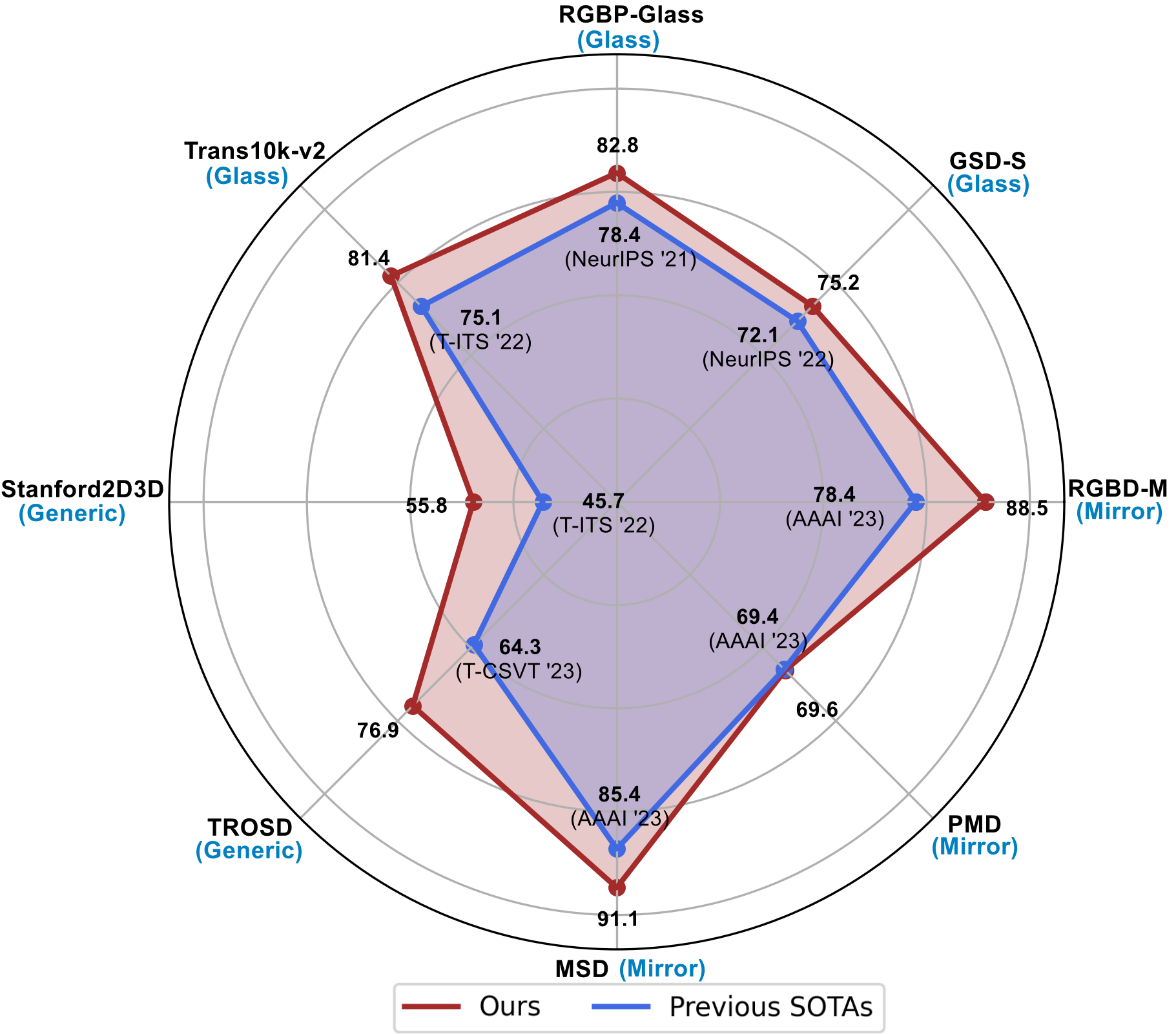}
    \caption{\small{
    Our method achieves competitive performance compared to previous methods across glass, mirror, and generic segmentation tasks. To maintain fairness, we only compare with methods that use the same input (only RGB image).}}
    \label{fig:all_datasets}
    \vspace{-6mm}
\end{figure}

\section{Introduction}
\label{sec:intro}
\vspace{-2mm}

Glass objects, such as windows, bottles, walls, and glass, have presented significant challenges to image segmentation due to their appearances being heavily influenced by the surrounding environment. Also, many robotic systems~\cite{xie2020segmenting,xie2021segmenting,zhang2022trans4trans} rely on sensor fusion techniques that use sonars or lidars, but these methods often struggle to detect transparent objects and misinterpret reflections as actual objects, leading to scan-matching issues. This is because transparent objects exhibit properties of refraction and reflection, which cause light to be reflected and to appear in surrounding areas, thereby misleading robot sensors and negatively impacting robot navigation, depth estimation, and 3D reconstruction. Hence, visual systems must deal with reflective surfaces, which would help them accurately identify glass barriers for effective collision prevention in workplaces, supermarkets, or hotels. Furthermore, in domestic and professional settings, visual systems should also be able to navigate fragile items such as vases and glasses. Therefore, a practical, robust, cost-effective, vision-based approach for transparent object segmentation is essential. However, current semantic segmentation algorithms~\cite{Xie2021SegFormerSA,setr,denseaspp,Chen2018eccv,ocnet} and even powerful foundation models, such as SAM~\cite{sam2023} and Semantic SAM~\cite{li2023semanticsam,chen2023semantic}, were not designed to address transparent and reflective objects, resulting in decreased performance in the presence of such objects. Importantly, the SAM model does not include semantics; in other words, it cannot yield masks with semantic information. The SAM model also presents the challenge of over-segmentation, thereby increasing the likelihood of false positives (see Supplementary for details).

Recent works on the human visual system~\cite{pnas2015human,jov2019perception,jov2022material} prove that \textbf{``humans rely on specular reflections and boundaries as key indicators of a transparent layer``}. 
Recent methods for segmenting transparent or glass objects have been proposed, along with various strategies to improve performance when dealing with glass objects. These strategies include focusing on the object's edges~\cite{xie2021segmenting,zhang2022trans4trans,trosd2023}, using depth information~\cite{trosd2023,emsanet}, analyzing reflections~\cite{Jiaying2021}, looking at how light polarizes~\cite{Xiang2021,PGSNet2022}, and utilizing the object's context or semantic information~\cite{gsds2022}. However, techniques that rely on polarization, depth, and semantic information~\cite{trosd2023,emsanet,Xiang2021,PGSNet2022,gsds2022} often require specialized equipment to collect data or extensive human labor to label it, which is inefficient. As far as we know, no method has combined both visual cues of boundary and reflection to improve segmentation performance. 

Therefore, we will focus on capturing the two visual cues into the segmentation models: boundary localization for shape inference and reflections for glass surface recognition. We introduce an \textit{efficient transformer-based architecture} tailored for segmenting {transparent} and {reflective} objects along with {general} objects. Then, our method captures the \textbf{glass boundaries} based on geometric cues and the \textbf{glass reflections} based on appearance cues within an enhanced feature module in our network. In doing so, we developed a Boundary Feature Enhancement (BFE) module to learn and integrate glass-boundary features to improve the localization and segmentation of glass-like regions. We supervise this module with a new boundary loss that uses the Sobel kernel to extract boundaries from the gradients of the predictions and the ground-truth objects' masks. Then, we introduce a Reflection Feature Enhancement (RFE) module that decomposes reflections into foreground and background layers, providing the network with additional features to distinguish glass-like from non-glass areas. By harnessing the power of transformer-based encoders and decoders, our framework can capture long-range contextual information, unlike previous methods that relied heavily on stacked attention layers~\cite{danet,yang2021capturing} or on combining CNN backbones with transformers~\cite{xie2021segmenting,setr,pvt2021}. These long-range visual cues are essential to reliably identify transparent objects, especially when they lack distinctive textures or share similar content with their surroundings~\cite{xie2021segmenting}. More importantly, we demonstrate that our method is robust to both transparent object segmentation and generic semantic segmentation tasks, with state-of-the-art performance for both scenarios across various datasets.

In summary, our contributions are as follows:
\begin{itemize}
    \item We introduce \textbf{TransCues}, an efficient transformer-based segmentation architecture that segments both transparent, reflective, and general objects.
    \item We propose the Boundary Feature Enhancement (BFE) module and a boundary loss that improves the accuracy of glass detection. 
    \item We present the Reflection Feature Enhancement (RFE) module, facilitating the differentiation between glass and non-glass regions.
    \item We conduct extensive experiments to demonstrate our method's competitive performance on diverse tasks, \eg semantic glass segmentation, glass and mirror segmentation, and generic semantic segmentation.
\end{itemize}

\begin{figure*}[t]
\centering
\includegraphics[width=0.96\linewidth]{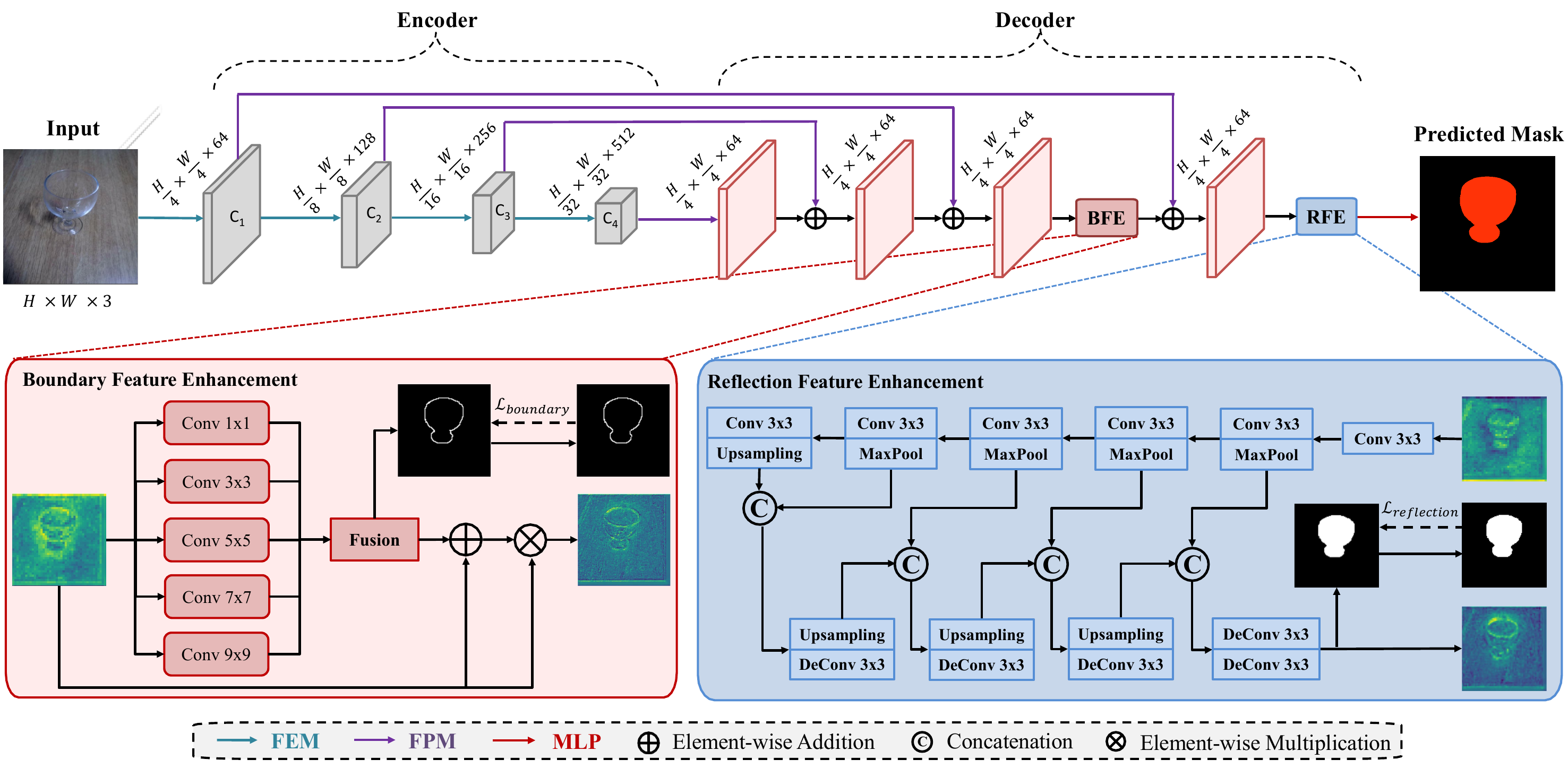}
\caption{\small{\textbf{Overview of our TransCues method.} An RGB image is processed by four FEM modules in the encoder for multi-scale feature extraction. These features are then refined by the decoder's FPM, BFE, and RFE modules, and ultimately converted into semantic labels via an MLP. Our main contributions, BFE and RFE modules, are elaborated in Sections~\ref{sec:BFE} and~\ref{sec:RFE}. 
}}
\label{fig:architecture}
\vspace{-2mm}
\end{figure*}

\section{Related Works}

\vspace{-1mm}
\subsection{Transparent Object Sensing and Segmentation}
\vspace{-1mm}
In the transparency settings, the color intensities of both glass and its background often match, making it challenging to differentiate between them. Traditional visual-aid systems, enhanced with ultrasonic sensors and RGB-D cameras, were developed to effectively identify transparent barriers like glass and windows~\cite{Zhiming2018}. On raw images, existing works have explored the use of transmission differences~\cite{Okazawa2019}, reflection cues~\cite{Jiaying2021}, and polarization~\cite{Xiang2021, PGSNet2022} for detecting transparency. Moreover, transparency segmentation methods~\cite{xie2021segmenting,He_2021_ICCV,mei_tpami_2023,qi2024glass,chang2024panoglassnet,chen2024toformer} address a range of objects, from opaque entities (windows and doors) to see-through items (cups and eyeglasses), focusing on discerning reflections and their boundaries to accurately detect and delineate transparent surfaces. Recently,~\cite{xie2021segmenting} introduced the Trans10K-v2 dataset, which prompts new research directions beyond conventional sensor fusion for transparent objects. This includes AdaptiveASPP~\cite{cao2021fakemix} for enhanced feature extraction and EBLNet~\cite{He_2021_ICCV} for improved global form representation. Furthermore, Trans4Trans~\cite{zhang2022trans4trans} is proposed to provide a lightweight general network for real-world applications. \textit{Building on these innovations}, our work aims to develop an efficient, robust, and transparent object segmentation solution suitable for general semantic segmentation and practical applications like robot navigation.

\vspace{-1mm}
\subsection{Mirror Segmentation}
\vspace{-1mm}
Closely related to glass segmentation is mirror segmentation, in which recent models have introduced high-level concepts to improve detection and localization~\cite{SANet2022,SATNet2023,VCNet2023,HetNet2023,wang2024shadow,xie2024csf}. For precise localization, SANet~\cite{SANet2022} utilizes the semantic relationships between mirrors and their surrounding environment. SATNet~\cite{SATNet2023} capitalizes on the natural symmetry between objects and their mirror reflections to accurately identify mirror locations. VCNet~\cite{VCNet2023}, in contrast, explores 'visual chirality'—a unique property of mirror images—and incorporates this through a specialized transformation process for effective mirror detection. Lastly, HetNet~\cite{HetNet2023} introduces a unique model that combines a contrastive module for initial mirror localization and a semantic logical reflection module for semantic analysis. \textit{Unlike existing works}, we consider mirror segmentation as a subproblem in glass segmentation that our proposed framework can also address effectively.

\vspace{-2mm}
\subsection{Transformer in Semantic Segmentation} 
\vspace{-1mm}
Since their introduction in natural language processing, transformers have been adopted and further investigated for computer vision tasks. One of the pioneers is Vision Transformer (ViT)~\cite{dosovitskiy2020vit}, which applies transformer layers to sequences of image patches. SETR~\cite{setr} and Segmenter~\cite{segmenter} are inspired by ViT and directly add upsampling and segmentation heads to learn long-range context from the initial layer. MaX-DeepLab~\cite{maxdeeplab}, and MaskFormer~\cite{cheng2021maskformer} study 2D image segmentation through the perspective of masked prediction and classification based on recent advances of object detection using transformers~\cite{detr}. As a result, several transformer-based methods for dense image segmentation have been developed~\cite{swin,Xie2021SegFormerSA}. Pyramid architectures for vision transformers have been proposed by~\cite{pvt2021,wang2021pvtv2} to capture hierarchical feature representations. Both ECANet~\cite{yang2021capturing} and CSWin transformer~\cite{Dong_2022_CVPR} recommend applying a self-attention mechanism in either vertical or horizontal stripes to gain advanced simulation capacity while minimizing computing overheads. NAT~\cite{Hassani2023NAT}, on the other hand, aims to simplify the standard attention mechanism, resulting in faster processing and reduced memory requirements. Recent methods have been trying to match the performance of transformer models using advanced CNN architectures. MogaNet~\cite{Li2022MogaNet} introduces two feature mixers with depthwise convolutions that efficiently process middle-order information across the spatial and channel spaces. InternImage~\cite{Wang2023Intern} utilizes deformable convolution, providing a large effective receptive field essential for tasks such as detection and segmentation, and offers adaptive spatial aggregation based on input and task-specifics. \textit{Collectively}, these approaches signify a shift towards more efficient, task-tailored CNN models that strive to replicate the success of transformers in various computer vision applications.

\vspace{-3mm}
\section{Our Proposed Method}
\vspace{-2mm}
Given an RGB image, defined as $\mathbf{I} \in \mathbb{R}^{H \times W \times 3}$, where $H$ and $W$ respectively denote the image height and width, glass segmentation aims to segment this image into semantic labels at each pixel, which can be expressed as $\mathbf{F} \in \mathbb{R}^{H \times W \times n_{class}}$, where $n_{class}$ represents the number of classes. While glass segmentation can typically be defined in the binary space, defining it along with glass, mirror, and non-glass objects makes for a semantic segmentation problem. Existing work has not considered boundary and reflective cues within the same framework. In particular, while boundary (edge) information was deemed to be captured in~\cite{xie2020segmenting,PGSNet2022,trosd2023} for glass segmentation, the reflective information of glass objects was not considered of high priority because of negligible reflections in the datasets. Meanwhile, mirror segmentation has been primarily analyzed for symmetric reflection to detect mirroring in the image~\cite{msd2019,pmd2020,HetNet2023,SATNet2023}.

\vspace{-1mm}
\subsection{Our TransCues} 
As depicted in Figure~\ref{fig:architecture}, our network aims to segment glasses in an image and assign their corresponding labels to each pixel. To handle variations in input image sizes across datasets, we standardize the resolution to either $512 \times 512$ or $768 \times 768$, ensuring consistent position-embedding dimensions throughout training and testing. Our network's architecture adheres to the well-established encoder-decoder structure, comprising the Feature Extraction Module (FEM), Feature Parsing Module (FPM), Boundary Feature Enhancement (BFE), and Reflection Feature Enhancement (RFE) modules. Specifically, the FEM module, based on the PVT architecture~\cite{pvt2021,wang2021pvtv2} in the encoder, efficiently captures multi-scale long-range dependencies in the input image. Drawing inspiration from~\cite{zhang2022trans4trans}, the FPM module offers a lightweight alternative to the FEM module, capturing detailed-to-abstract representations of transparent objects across $C_1$ to $C_4$. The details of FEM and FPM are provided in the Supplementary.

To enrich the feature learning capabilities of our network, we propose capturing the glass boundaries with a \textit{geometric cue} (BFE module) and the glass reflections with an \textit{appearance cue} (RFE module) to differentiate glass from non-glass regions. 
Boundaries often present as high-contrast edges around transparent objects, a characteristic that aligns well with human visual perception. In detail, we employ the BFE module to amplify the boundary characteristics inherent in transparent features. This enhancement of boundary cues facilitates more accurate segmentation of transparent objects.

Alternatively, reflections on glass surfaces may not always be prominent, making it challenging to design glass segmentation methods. Following this, we feed the boundary-enhanced features into the RFE module. This step is crucial because while most transparent objects exhibit reflections, not all reflective objects are transparent. The RFE module thus plays a pivotal role in distinguishing between these two categories. These cues enhance our network's ability to capture fine details and long-range contextual information for transparent features. 

\begin{figure}[t]
\centering
\includegraphics[width=0.98\linewidth]{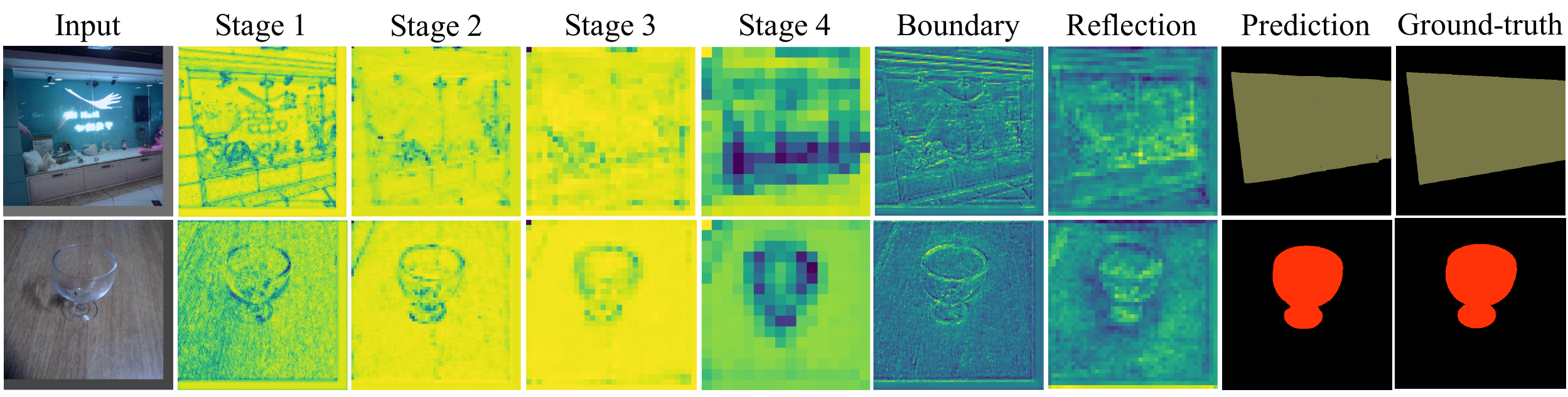}
\caption{\small{Visualization of feature maps of our method. Zoom in for better visualization.}}
\label{fig:vis_features}
\vspace{-4mm}
\end{figure}

Consequently, as image data progresses through our decoder, it integrates contextual information of varying resolutions, preserving the fine-grained information of transparent and reflective features. By systematically addressing both boundary and reflection cues, our network achieves a nuanced and effective approach to segmenting these challenging object types. Finally, a compact MLP layer is employed to predict semantic labels for each pixel. As shown in Figure~\ref{fig:vis_features}, throughout each stage of the encoder and the boundary and reflection modules, our feature maps clearly distinguish the glass region, including its defining boundary and reflection, from non-glass surfaces. The following sections will discuss the BFE and RFE modules in more detail.

\vspace{-1mm}
\subsection{Boundary Feature Enhancement Module} 
\label{sec:BFE}
\vspace{-1mm}
Inspired by human perception, incorporating boundary information can significantly benefit segmentation and localization tasks involving glass recognition~\cite{He_2021_ICCV,mei_tpami_2023}. To implement this concept, the BFE module is based on the ASPP module~\cite{Chen2018eccv,mei_tpami_2023}, yet it is more specialized toward identifying and integrating boundary characteristics of glass into our transformer architecture. Contrary to the approach in~\cite{xie2020segmenting}, which uses an additional boundary stream (an encoder-decoder branch) for boundary feature extraction and integration with primary-stream features, our BFE module is more streamlined. It derives boundary features directly from the targeted input features, bypassing the necessity for an additional stream. As shown in Figure~\ref{fig:architecture}, the BFE module is designed to enhance feature learning before the last layer of our decoder so that the reflection module can subsequently improve the features in the next step. We empirically found that this placement of the BFE module achieves better performance and reduced memory usage compared to placing BFE earlier in the decoder (please see Section~\ref{sec:tos_ablation} for more details).

The BFE begins by taking input features $\mathcal{X}_0$. These features are then processed through four parallel blocks, each dedicated to extracting multi-scale boundary features $\mathcal{F}_i(.)$ for $i=1,2,3,4,5$. Within each block, a convolution layer ($C(.)$) with different kernels and paddings is followed by batch normalization ($BN(.)$) and ReLU activation ($ReLU(.)$) operations, resulting in $\mathcal{F}i = ReLU(BN(C(X)))$. These multi-scale boundary features are subsequently fused using the Fusion module ($\mathcal{F}_{\text{fuse}} = C(\mathcal{F}_1 + \mathcal{F}_2 + \mathcal{F}_3 + \mathcal{F}_4+ \mathcal{F}_5)$), effectively aggregating shape properties and forming the glass boundary features. The output of the Fusion module then undergoes a convolutional layer to predict the boundary map, supervised by the Boundary loss. Finally, the enhanced boundary features $\mathcal{X}_e$ are obtained by aggregating the output of the Fusion module with the input features to locate glass regions, especially their boundaries, as expressed by the following equation:
\begin{equation}
    \mathcal{X}_e = (\mathcal{F}_{\text{fuse}}(\mathcal{F}_i(\mathcal{X}_0)) + \mathcal{X}_0) \times \mathcal{X}_0
\end{equation}
where $+$ and $\times$ denote element-wise addition and multiplication, respectively.

\noindent\textbf{Boundary loss.} 
The Sobel kernel, also known as the Sobel-Feldman filter, is widely used in image processing and computer vision, primarily for edge detection. It highlights image boundaries by analyzing the 2D gradient and emphasizing high-frequency regions. Our Boundary loss ($\mathcal{L}_{b}$) leverages the Sobel filter to measure how closely the gradients of a predicted mask match those of the ground truth mask, employing the Dice loss~\cite{dice2016}:
\begin{equation} 
    \mathcal{L}_{b} = \mathrm{dice}(\nabla_x{\hat{M}} \oplus \nabla_y{\hat{M}}, \nabla_x{M_{GT}} \oplus \nabla_y{M_{GT}})
\end{equation} 
where $\hat{M}$ is predicted object mask and $M_{GT}$ is ground truth object mask. $\nabla_x$ and $\nabla_y$ denote the gradient along $x$-axis and $y$-axis computed by the Sobel filter. $\oplus$ represents the combination of the gradient maps into a single feature map. In our implementation, we define the combination $\oplus$ by: 
\vspace{-2mm}
\begin{equation} 
    a \oplus b = \max\left(\frac{1}{2} (a + b), \tau\right)
\end{equation} 
where $\tau$ is set to $0.01$ to reduce noise in the gradient maps.

\vspace{-1mm}
\subsection{Reflection Feature Enhancement Module}
\label{sec:RFE}
\vspace{-1mm}
To enhance the recognition of glass surfaces, we introduce the Reflection Feature Enhancement (RFE) module, which capitalizes on the high reflectivity of glass when illuminated. These reflections provide valuable cues for recognizing glass surfaces in images~\cite{msd2019,rgbdm2021}. Note that if the reflection on the glass surface is insufficient to be discerned by our RFE module, our model may struggle to accurately detect the glass surface. In this scenario, correctly identifying glass surfaces is challenging, even for humans. Please check the Supplementary for discussion and analysis on the need for the RFE module.

\textit{In our design}, the RFE module is placed after the last layer of the decoder, after the boundary feature enhancement module (please check Section~\ref{sec:tos_ablation} for more detail). The RFE module employs a sophisticated convolution-deconvolution architecture~\cite{YuKoltun2016}, which takes input features $Y$ and produces an enhanced feature map $Y_{e}$. This architecture allows the module to capture and process information at multiple levels of abstraction, which is essential for handling complex visual cues like reflections. Unlike the other reflection removal models~\cite {Zhang_2018_CVPR,Dong2021Removal,Song2023removal} that primarily address global reflections (assuming the entire input image is covered by glass), our RFE module targets detecting local reflections to locate glass surfaces. 

\textit{In detail}, the encoder network $\mathcal{E}$ is responsible for extracting relevant features from the input. It consists of five blocks, each composed of a convolutional layer followed by batch normalization, ReLU activation, and either a Max-Pooling $\mathcal{P}_{max}(.)$ or Upsampling layer $\mathcal{P}_{up}(.)$. Each encoder block can be defined as follows:
\begin{equation} 
\mathcal{E}_i = \mathcal{P}^i\left(\text{ReLU}\left(\text{BN}\left(\text{C}(\mathcal{E}_{i-1})\right)\right)\right), \quad i \in [1..5]
\end{equation}
where $\mathcal{E}_0 = C(Y)$, $\mathcal{P}^i$ is the Max-Pooling or Upsampling layer and when $i=5$, $\mathcal{P}^i$ will be $\mathcal{P}_{Up}(.)$ instead of $\mathcal{P}_{max}(\cdot)$. 

\textit{Consequently}, the decoder network $\mathcal{D}$ works in conjunction with the encoder to reconstruct and enhance features. It also comprises four blocks, interconnected by an Upsampling layer $\mathcal{P}_{up}(.)$ or a Deconvolutional layer $DC(\cdot)$, along with batch normalization and ReLU activation. Notably, the output of the preceding decoder block and the corresponding feature map $e_i = \mathcal{E}_i$ from the encoder block are concatenated before being fed into the subsequent decoder block. This facilitates the seamless flow of information across the network, enhancing its ability to capture and retain essential features of the reflective areas. Each decoder block can be formulated as follows: 
\newcommand{\concat}{\mathbin{\smallfrown}}
\begin{equation}
\mathcal{D}_j = \mathcal{P}^j(\text{ReLU}(\text{BN}(\text{DC}(\mathcal{D}_{j-1} \frown \mathcal{E}_{5-j})))), \quad j \in [1..4] 
\end{equation}
where $\frown$ is concatentation operation, $\mathcal{D}_0 = \mathcal{E}_5$, $\mathcal{P}^j$ is the Upsampling or Deconvolutional layer and when $j=4$, $\mathcal{P}^j$ will be $DC(\cdot)$ instead of $\mathcal{P}_{Up}(\cdot)$.

The decoder network's output is split into two tensors: the first tensor represents reflection mask $M_{\text{rf}}$, utilized for optimizing the reflection loss, while the second tensor contains the enhanced reflective features $Y_{e}$, which have been processed to capture and emphasize reflection information. 

\vspace{-1mm}
\subsection{Loss Functions} 
\vspace{-1mm}
We use the softmax cross-entropy loss as our semantic loss $\mathcal{L}_{s}$ for supervising the semantic mask prediction and the ground truth (GT) semantic mask:
\begin{equation}
    \mathcal{L}_{s} = \mathrm{ce}(\hat{M}, M_{GT})
\end{equation} 
where $\mathrm{ce}(.)$ is the softmax cross-entropy loss. 

To supervise reflection, we also use the softmax cross-entropy loss for our reflection loss $\mathcal{L}_{r}$. However, there is no GT for the reflection mask, and we assume pseudo GT for the reflection mask to span common categories with reflective appearance, such as window, door, cup, bottle, \etc Therefore, we extract pseudo GT with the reflective appearance in the GT semantic map $M_{GT}$. Note that, as our pseudo GT may contain opaque appearances, we empirically found that this noise is not severe enough to affect the performance of the RFE module. The reflection loss is:
\begin{equation}
    \mathcal{L}_{r} = \mathrm{ce}(M_{\text{rf}}, \phi(M_{GT}))
\end{equation} 
where $\phi(.)$ is a function to extract pseudo GT with reflective appearance in the GT semantic map $M_{GT}$. 

\noindent The total loss for our training is:
\begin{equation}
    \label{eq:tos_final_loss}
    \mathcal{L} = \alpha \mathcal{L}_{s} + \beta \mathcal{L}_{b} + \gamma \mathcal{L}_{r}
\end{equation}
where $\alpha$, $\beta$ and $\gamma$ are hyper-parameters and are empirically set as [1.0,0.1,0.1] according to the experimental results.

\vspace{-1mm}
\section{Experiments}
\vspace{-1mm}
\noindent\textbf{Datasets.} We comprehensively evaluated our proposed method on diverse datasets to demonstrate its exceptional performance and versatility. For details of the datasets, method implementations, experiments, and further analyses, please refer to the supplementary material.

\vspace{-1mm}
\subsection{Qualitative and Quantitative Results}
\vspace{-1mm}

We evaluated our method's performance across three distinct tasks: glass, mirror, and generic segmentation. 
To ensure \textbf{fair comparisons}, we have carefully \textbf{selected our model variants} (Ours-X with X is postfixes: -T, -S, -M, -L, -B1, -B2, -B3, -B4, and -B5, represented the model's size as PVTv1 Tiny, Small, Medium, Large, and PVTv2 B1-5, respectively) that have \textbf{similar model's size or complexity} used by other methods, as indicated in the respective tables. 

\vspace{1mm}
\noindent\textbf{Glass Segmentation.} 
We benchmarked our method against recent glass segmentation methods on binary and semantic segmentation tasks. For the binary glass segmentation task, as shown in Table~\ref{tab:binary_glass}, our method (Ours-B4) achieves the pinnacle of mIoU(\%) scores, outpacing all other competing methods, which include (SegFormer~\citep{Xie2021SegFormerSA}, GSD~\citep{Jiaying2021}, SETR~\citep{setr}, GDNet~\citep{Mei_2020_CVPR}). Specifically, it surpasses the runner-up method by margins of 4.35\% for RGB-P and 2\% for GSD-S. Noteworthy is our method's balance of performance and computational efficiency, which registers relatively lower GFLOPs than its peers. Shifting our focus to the semantic glass segmentation task, where the challenge extends beyond merely detecting glass areas to classifying them into 11 fine-grained categories, our method still reigns supreme. It surpasses competing approaches such as  (Trans4Trans~\citep{zhang2022trans4trans}, DenseASPP~\citep{denseaspp}, DeepLabv3+~\citep{Chen2018eccv}, OCNet~\citep{ocnet}, Trans2Seg~\citep{xie2021segmenting}) by a substantial 4.15\% margin in terms of mIoU performance. This dominance in accuracy does not come at the expense of efficiency, as evidenced in Table~\ref{tab:sem_glass}. These comprehensive evaluations underscore the effectiveness of our approach across diverse glass segmentation scenarios, affirming its position as a top-performing and computationally efficient choice for these tasks. 

\begin{figure*}[!htb]
    \centering
    \includegraphics[width=0.96\linewidth]{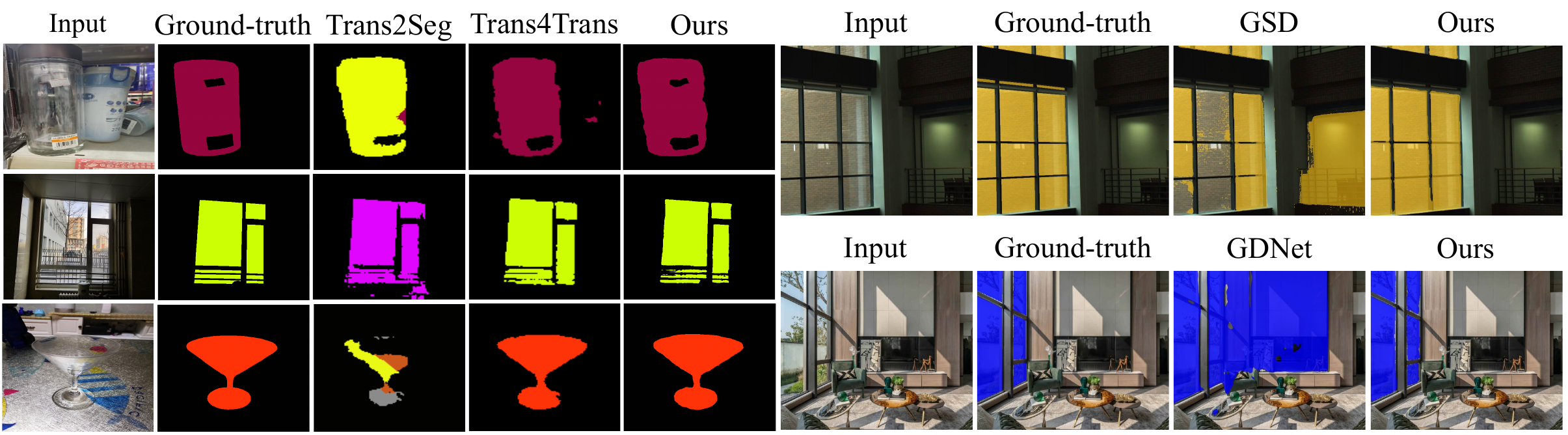}
    \caption{Comparison of glass segmentation methods on Trans10K-v2 (left), RGB-P (top-right), and GSD-S (bottom-right) datasets.}
    \label{fig:vis_glass}
    \vspace{-2mm}
\end{figure*}

As shown in Figure~\ref{fig:vis_glass}, we observe that recent approaches, such as GDNet and Trans2Seg, may over-detect glass regions in certain images but under-detect them in others, such as GSD. In contrast, our method can accurately identify glass portions of diverse dimensions and morphologies, effectively differentiating them from look-alike non-glass regions in complex images (such as the one showcased in the top right), thanks to the BFE and RFE modules, which leverage boundary and reflection cues, helping our method perform better in challenging scenarios.

\begin{table}[!t]
\centering
\caption{\small Binary Glass Segmentation on RGB-P, GSD-S. 
We reported mIoU(\%) for both datasets. 
}
\vspace{-2mm}
\resizebox{0.96\linewidth}{!}{%
\begin{tabular}{ll|ccc}
\toprule
\textbf{\textbf{Method}}            & \textbf{\textbf{Backbone}} & \textbf{\textbf{GFLOPs}~$\downarrow$} & \textbf{RGB-P} & \textbf{GSD-S} \\ \midrule
SegFormer & MiT-B5                     & 70.2                                 & 78.4          & 54.7          \\
\rowcolor{gray!15} Ours-B4                             & PVTv2-B4                   & 79.3                                & \textbf{82.1} & \textbf{74.1} \\ 
GSD              & ResNeXt-101                & 92.7                                 & 78.1          & 72.1          \\
SETR                    & ViT-Large                  & 240.1                                & 77.6          & 56.7          \\
GDNet          & ResNeXt-101                & 271.5                                & 77.6          & 52.9          \\
\bottomrule
\end{tabular}
}
\label{tab:binary_glass}
\end{table}

\begin{table}[!t]
\centering
\caption{\small Semantic Glass Segmentation on Trans10K-v2.}
\vspace{-2mm}
\resizebox{0.9\linewidth}{!}{%
\begin{tabular}{l|ccc}
\toprule
\textbf{Method} & \textbf{GFLOPs}~$\downarrow$ & \textbf{Accuracy}~$\uparrow$ & \textbf{mIoU}~$\uparrow$ \\ 
\midrule \midrule
Trans4Trans-M &  34.38 &  95.01  & 75.14 \\ 
DenseASPP & 36.20 & 90.86 & 63.01 \\
\rowcolor{gray!15} Ours-B2 &  37.03 &  \textbf{95.92}  & \textbf{79.29} \\ 
DeepLabv3+ & 37.98 & 92.75 & 68.87 \\
OCNet  & 43.31 & 92.03 & 66.31 \\
Trans2Seg &  49.03 &  94.14  & 72.15 \\ 
\bottomrule
\end{tabular}
}
\vspace{-2mm}
\label{tab:sem_glass}
\end{table}

\begin{figure}[!t]
    \centering
    \includegraphics[width=0.94\linewidth]{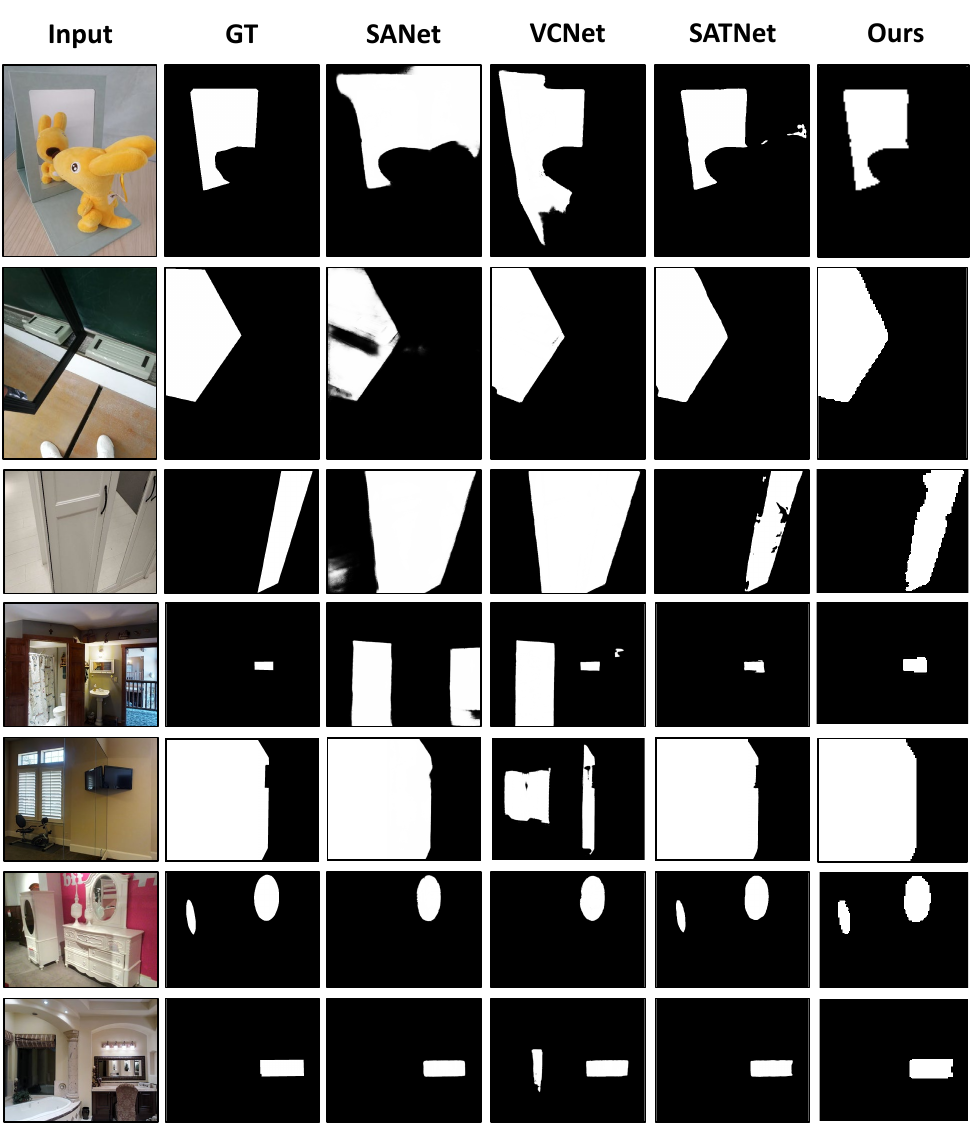}
    \caption{\small{Qualitative comparison of our method with other methods on MSD, PMD, and RGBD-Mirror datasets.}}
    \label{fig:mirror_full}
    \vspace{-3mm}
\end{figure}

\vspace{1mm}
\noindent\textbf{Mirror Segmentation.}
To demonstrate the robustness of our approach for reflective surfaces, we rigorously evaluated our method on three standard binary mirror segmentation datasets: MSD, PMD, and RGBD-Mirror. These datasets are chosen for their resemblance to the reflective characteristics inherent in glass objects. Ours-B3 was chosen as a representative model for a balanced and comparable evaluation. We compared its performance against recent state-of-the-art methodologies, such as  SANet~\citep{SANet2022}, VCNet~\citep{VCNet2023}, and SATNet~\citep{SATNet2023}. The outcomes, detailed in Table~\ref{tab:mirror} and Figure~\ref{fig:mirror_full}, unequivocally highlight our method's supremacy in mirror segmentation, surpassing the competition across various metrics.

\begin{table*}[!htb]
\centering
\caption{Binary Mirror Segmentation on MSD, PMD, and RGBD-Mirror datasets. Our method achieves the best performance in terms \\ of all the evaluation metrics. }
\vspace{-2mm}
\resizebox{0.88\linewidth}{!}{%
\begin{tabular}{cc|ccc|ccc|ccc}
\toprule
\multirow{2}{*}{\textbf{Method}} & \multirow{2}{*}{\textbf{Backbone}} & \multicolumn{3}{c|}{\textbf{MSD}}                                                   & \multicolumn{3}{c|}{\textbf{PMD}}                                                   & \multicolumn{3}{c}{\textbf{RGBD-Mirror}}                                                 \\ \cmidrule(l){3-11} 
                                 &                                    & \textbf{IoU}~$\uparrow$ & $\mathbf{F_\beta}$~$\uparrow$ & \textbf{MAE}~$\downarrow$ & \textbf{IoU}~$\uparrow$ & $\mathbf{F_\beta}$~$\uparrow$ & \textbf{MAE}~$\downarrow$ & \textbf{IoU}~$\uparrow$ & $\mathbf{F_\beta}$~$\uparrow$ & \textbf{MAE}~$\downarrow$ \\ \midrule
SANet~\citep{SANet2022}                            & ResNeXt101                         & 79.85                   & 0.879                         & 0.054                     & 66.84                   & 0.837                         & 0.032                     & 74.99                   & 0.873                         & 0.048                     \\
VCNet~\citep{VCNet2023}                            & ResNeXt101                         & 80.08                   & 0.898                         & 0.044                     & 64.02                   & 0.815                         & 0.028                     & 73.01                   & 0.849                         & 0.052                     \\
SATNet~\citep{SATNet2023}                           & Swin-S                             & 85.41                   & 0.922                         & 0.033                     & 69.38                   & 0.847                         & 0.025                     & 78.42                   & 0.906                         & 0.031                     \\
\rowcolor{gray!15} Ours-B3       & PVTv2-B3                           & \textbf{91.04}          & \textbf{0.953}                         & \textbf{0.028}                     & \textbf{69.61}          & \textbf{0.853}                         & \textbf{0.021}                     & \textbf{88.52}          & \textbf{0.954}                         & \textbf{0.027}                     \\ \bottomrule
\end{tabular}%
}
\vspace{-2mm}
\label{tab:mirror}
\end{table*}

\begin{figure}[!t]
    \centering
    \includegraphics[width=0.98\linewidth]{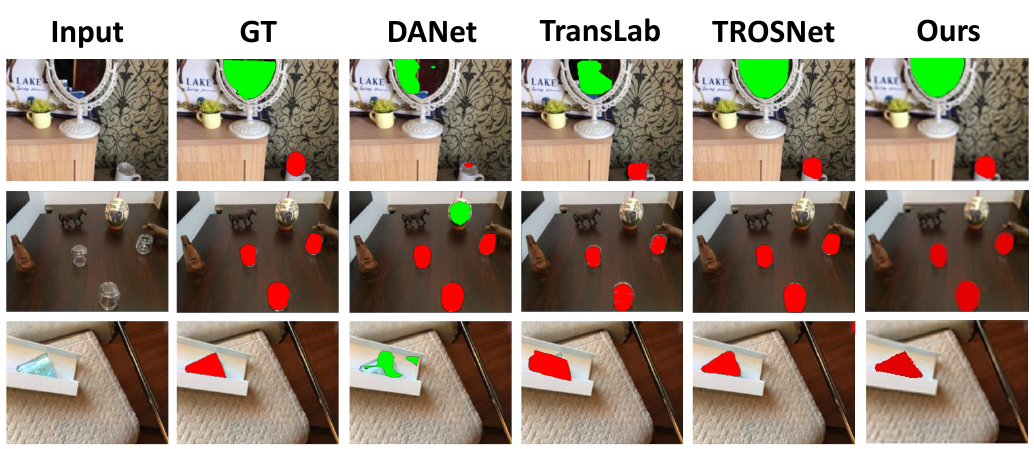}
    \caption{\small{Qualitative comparison of our method with other methods on the TROSD dataset.}}
    \label{fig:trosd}
    \vspace{-3mm}
\end{figure}

\vspace{1mm}
\noindent\textbf{Generic Segmentation.}
To evaluate our method, we compared it with existing approaches (Trans4Trans~\citep{zhang2022trans4trans}, TransLab~\citep{xie2020segmenting}, DANet~\citep{danet}, TROSNet~\citep{trosd2023}, Trans2Seg~\citep{xie2021segmenting}, PVT~\citep{pvt2021}). As detailed in Table~\ref{tab:trosd} and Figure~\ref{fig:trosd}, our method outshines SOTA competitors on the TROSD dataset (a dedicated dataset for transparent and reflective object segmentation), underscoring its effectiveness in handling complex transparent and reflective object segmentation. This is mainly because our approach focuses on low-level features, enabling accurate identification and preservation of content differences along the borders of transparent and reflective objects. Moreover, to demonstrate our generalization ability, we evaluate our method on the large-scale real-world Stanford2D3D dataset, which comprises common and transparent objects (less than 1\% of the total images). In Table~\ref{tab:s2d3d}, our method outperforms other existing works (about $8.25\%$ better performance in mIOU) in semantic scene segmentation, demonstrating its robustness in discerning specific objects' appearance and normal scenes.


\begin{table}[!htb]
\centering
\caption{\small Comparison of various methods on TROSD dataset. R: reflective objects. T: transparent objects. B: background.
}
\vspace{-2mm}
\resizebox{0.99\linewidth}{!}{%
\begin{tabular}{lHc|ccccc}
\toprule
\multirow{2}{*}{\textbf{Method}} & \multirow{2}{*}{\textbf{Input}} & \multirow{2}{*}{\textbf{Backbone}} & \multicolumn{3}{c}{\textbf{IOU~$\uparrow$}} & \multirow{2}{*}{\textbf{mloU~$\uparrow$}} & \multirow{2}{*}{\textbf{mAcc~$\uparrow$}} \\ 
                                 &                                 &                                    & \textbf{R} & \textbf{T} & \textbf{B} &                                    &                                    \\ 
\midrule \midrule
TransLab                         & RGB                             & ResNet-50                          & 42.57      & 50.72      & 96.01      & 63.11                              & 68.72                              \\
DANet                            & RGB                             & ResNet-101                         & 42.76      & 54.39      & 95.88      & 64.34                              & 70.95                              \\
TROSNet                          & RGB                             & ResNet-50                          & 48.75      & 48.56      & 95.49      & 64.26                              & 75.93                              \\
\rowcolor{gray!15} Ours-B3                          & RGB                             & PVTv2                                & \textbf{67.25}      & \textbf{67.23}      & \textbf{97.69}      & \textbf{77.39}                              & \textbf{87.62}                               \\
\bottomrule
\end{tabular}%
}
\vspace{-2mm}
\label{tab:trosd}
\end{table}

\begin{table}[!htb]
\centering
\caption{\small Comparison with state-of-the-art methods on Stanford2D3D dataset.
}
\vspace{-2mm}
\resizebox{0.88\linewidth}{!}{%
\begin{tabular}{l|ccc}
\toprule
\textbf{Method} &  \textbf{GFLOPs}~$\downarrow$ & \textbf{MParams}~$\downarrow$ & \textbf{mIoU}~$\uparrow$ \\ 
\midrule \midrule
Tran4Tran-M & 34.38  & 43.65 & 45.73 \\
\rowcolor{gray!15} Ours-B2 &  37.03 & 27.59 & \textbf{53.98} \\
Trans2Seg-M &  40.98 & 30.53 & 43.83 \\
PVT-M & 49.00  & 56.20 & 42.49 \\
\bottomrule
\end{tabular}
}
\vspace{-2mm}
\label{tab:s2d3d}
\end{table}

\begin{figure}[!t]
    \centering
    \includegraphics[width=0.98\linewidth]{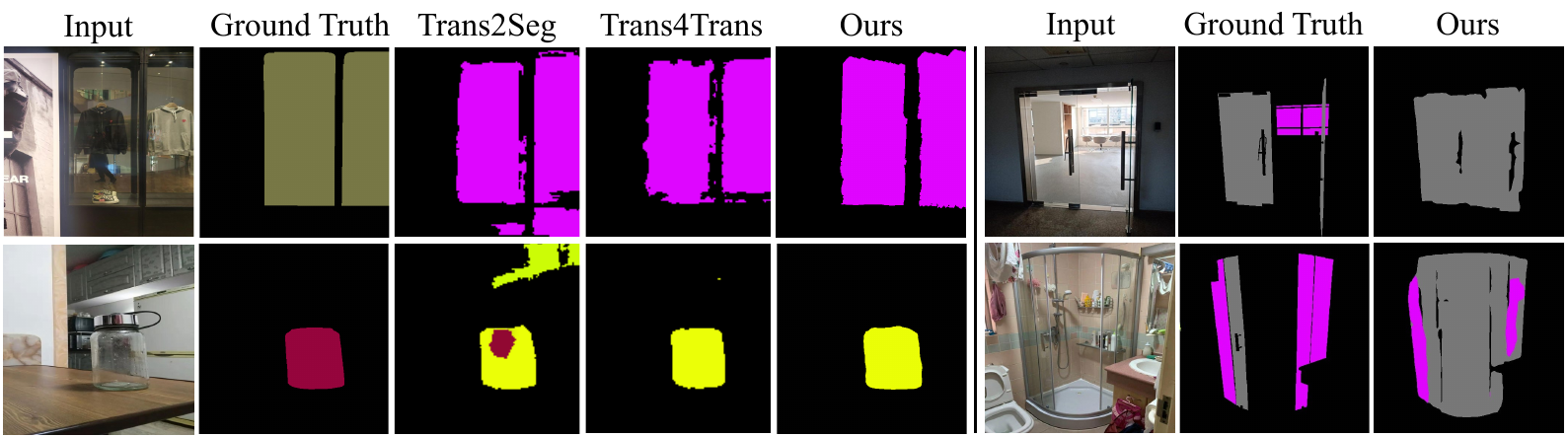}
    \caption{\small{Failure cases of our method and existing methods on the Trans10K-v2 dataset.}}
    \label{fig:tos_failure}
    \vspace{-3mm}
\end{figure}

\vspace{1mm}
\noindent\textbf{Failure Cases.} Figure~\ref{fig:tos_failure}--left shows failure cases of our method and others on Trans10K-v2. Our method would confuse and fail to segment the object with properties similar to those of others. In such a scenario, even humans would struggle to differentiate between these transparent things. 
Despite assigning wrong label, our method can still maintain the object's shape. 
We also show several failure instances (Figure~\ref{fig:tos_failure}--right) in our system that misinterpret non-glass areas as glass because they seem and behave the same, \eg the door frame with reflection and distortion.

\vspace{-1mm}
\subsection{Ablation studies}
\label{sec:tos_ablation}

We present ablation studies to verify various aspects of our model's design and underscore the significance of each module. Any alterations or omissions to the proposed design led to noticeable performance drops, which justifies our choice of the transformer architecture and the boundary and reflection feature learning components.

\vspace{1mm}
\noindent\textbf{Effectiveness of different modules.} 
To assess the contribution of both the proposed BFE and RFE modules to our architecture's performance, we systematically evaluated the model under various configurations: \ding{182} \textbf{Baseline Model} (PVTv1-T or PVTv2-B1 without BFE and RFE): this served as our control group, where both the BFE and RFE modules were excluded. Results indicate a foundational performance against which the other configurations could be compared. \ding{183} \textbf{Incorporation of BFE:} When only the BFE module was integrated into our network, we noticed a significant performance enhancement. However, this performance did not reach the potential of the combined BFE and RFE configuration. This proved that while BFE is essential, it is most effective in tandem with RFE. \ding{184} \textbf{Incorporation of RFE:} Similarly, adding only the RFE module to the baseline network also increased performance. This emphasized the value of detecting reflections in transparent objects for the segmentation task. \ding{185} \textbf{Combined Integration of BFE and RFE:} both modules were simultaneously integrated into our network. The performance gain observed in this configuration, as shown in Table~\ref{tab:v2_bfe_rfe}, was the most pronounced, with gains of \textbf{6.36\%} and \textbf{7.61\%} in mIoU on the Trans10K-v2 and Stanford2D3D datasets, respectively. This confirms that the combined effects of boundary and reflection cues significantly augment the network's segmentation capabilities.

Interestingly, the ablation studies further explain why our method performs well for generic segmentation, as demonstrated on the Stanford2D3D dataset. Table~\ref{tab:v2_bfe_rfe} shows that the boundary module yields the largest performance gain compared to the reflection module. This means that, for generic segmentation, where reflection feature learning yields negligible improvement, our boundary feature learning remains effective across general semantic labels.

\begin{table}[!t]
\centering
\footnotesize
\caption{\small{Effectiveness of different modules of our method on Trans10K-v2 dataset~\cite{xie2021segmenting} and Stanford2D3D (S2D3D) dataset~\cite{s2d3d}. We reported mIoU(\%) as a metric in this study. The last row corresponds to our method (Ours-B1).
}}
\setlength{\tabcolsep}{3pt}
\renewcommand{\arraystretch}{1.1}
\resizebox{0.49\textwidth}{!}{
\begin{tabular}{lcccc|ll}
\toprule
\textbf{Backbone} &  \textbf{FLOPs} &  \textbf{Params} &  \textbf{BFE} & \textbf{RFE} & \textbf{S2D3D} & \textbf{Trans10K} \\ \midrule \midrule
PVTv1-T  &  10.16          & 13.11          &  -          & -          & 45.19 & 69.44 \\
PVTv2-B1 &  11.48          & 13.89          &  -          & -          & 46.79 {\scriptsize \textcolor{teal}{+1.6}} & 70.49 {\scriptsize \textcolor{teal}{+1.05}} \\
PVTv2-B1 &  13.22          & 14.37          &  -          & \ding{51} & 48.12 {\scriptsize \textcolor{teal}{+2.93}} & 72.65 {\scriptsize \textcolor{teal}{+3.21}} \\
PVTv2-B1 &  19.55          & 14.39          &  \ding{51} & -          & 50.22 {\scriptsize \textcolor{teal}{+5.03}} & 74.89 {\scriptsize \textcolor{teal}{+5.45}} \\
PVTv2-B1 &  21.29          & 14.87          &  \ding{51} & \ding{51} & \textbf{51.55} {\scriptsize \textcolor{teal}{+6.36}} & \textbf{77.05} {\scriptsize \textcolor{teal}{+7.61}}\\
\bottomrule
\end{tabular}
}
\label{tab:v2_bfe_rfe}
\vspace{-1mm}
\end{table}

\vspace{1mm}
\noindent\textbf{Placement of modules.}
Incorporating both visual cues into the same framework is nontrivial, as certain features may be best captured at different stages. In our framework, we find the related features better captured at the end of the Decoder layers, with BFE followed by the RFE module. Table~\ref{tab:placement} shows that the aforementioned modules' positions and processing order matter.

\begin{table}[!t]
\footnotesize
\setlength{\tabcolsep}{5pt}
\renewcommand{\arraystretch}{1.1}
\centering
\caption{\small{Placement of modules on Trans10k-v2 dataset. $X \rightarrow Y$ and $X \parallel Y$ means placing $X$ before $Y$ and placing them in parallel and concatenate their outputs in Figure~\ref{fig:architecture}.
}}
\vspace{-1mm}
\resizebox{0.46\textwidth}{!}{%
\begin{tabular}{l|Hcc}
\toprule
\multicolumn{1}{l|}{\textbf{Variants}}  & \textbf{FLOPs} & \textbf{MParams~$\downarrow$} & \textbf{mIoU~$\uparrow$} \\ \midrule \midrule
Baseline & 11.48 & 13.89 & 70.49 \\
\qquad \qquad + \, RFE $\rightarrow$ BFE in Encoder & 12.98 & 48.98 & 73.54 \\
\qquad \qquad + \, BFE $\rightarrow$ RFE in Encoder & 20.25 & 48.99 & 74.12 \\
\qquad \qquad + \, RFE $\rightarrow$ BFE in Decoder & 22.60 & 14.90 & 75.11 \\
\qquad \qquad + \, BFE $\parallel$ RFE in Decoder & 22.61 & 14.91 & 75.44 \\
\midrule
TransCues (BFE $\rightarrow$ RFE in Decoder) & 21.29 & 14.87 & \textbf{77.05} \\ \bottomrule
\end{tabular}%
}
\label{tab:placement}
\vspace{-3mm}
\end{table}

\section{Conclusions}

In conclusion, this work proposes a method to segment transparent, opaque, and general objects using a pyramidal transformer architecture. Our method exploits two important visual cues, boundary and reflection features, which significantly improve performance in both transparent and generic segmentation tasks. We extensively evaluated our proposed method on several benchmark datasets, demonstrating its robustness in various scenarios. Our architecture is a fully transformer-based method built upon the PVT~\cite{pvt2021}. Therefore, some limitations remain that reduce our method's capabilities for visual tasks. Firstly, the position encoding in our network is fixed-size, requiring a resizing step that can damage and distort the object's shape. Secondly, similar to other vision transformer-based methods, our network incurs a relatively high computational cost when processing high-resolution images. Finally, as stated earlier, we use the same positional embedding as ViT~\cite{dosovitskiy2020vit} and PVT~\cite{pvt2021}, which is insufficient for images of arbitrary resolution. 
In future work, we would like to investigate how to address the above limitations and improve failure cases. In addition, we would like to investigate the extension of our method to other modalities, including depth images, event data, videos, and dynamic scenes. 

\section*{APPENDIX}

\begin{abstract}
Our supplementary material has three sections. Section~\ref{app-sec:arch} shows the detailed architecture of each module in our proposed method and provides hyperparameters of each scale of PVTv1 and PVTv2. Section~\ref{app-sec:setup} contains the experimental setup, including datasets, implementation details, and evaluation metrics. Additional analysis is detailed in Section~\ref{app-sec:exp}, together with quantitative and qualitative results of each dataset. In addition, we provide images and videos as a demo of deploying our method on a real robot.
\end{abstract}

\section{Network architecture}
\label{app-sec:arch}

\begin{figure*}[!t]
\centering
\includegraphics[width=0.98\linewidth]{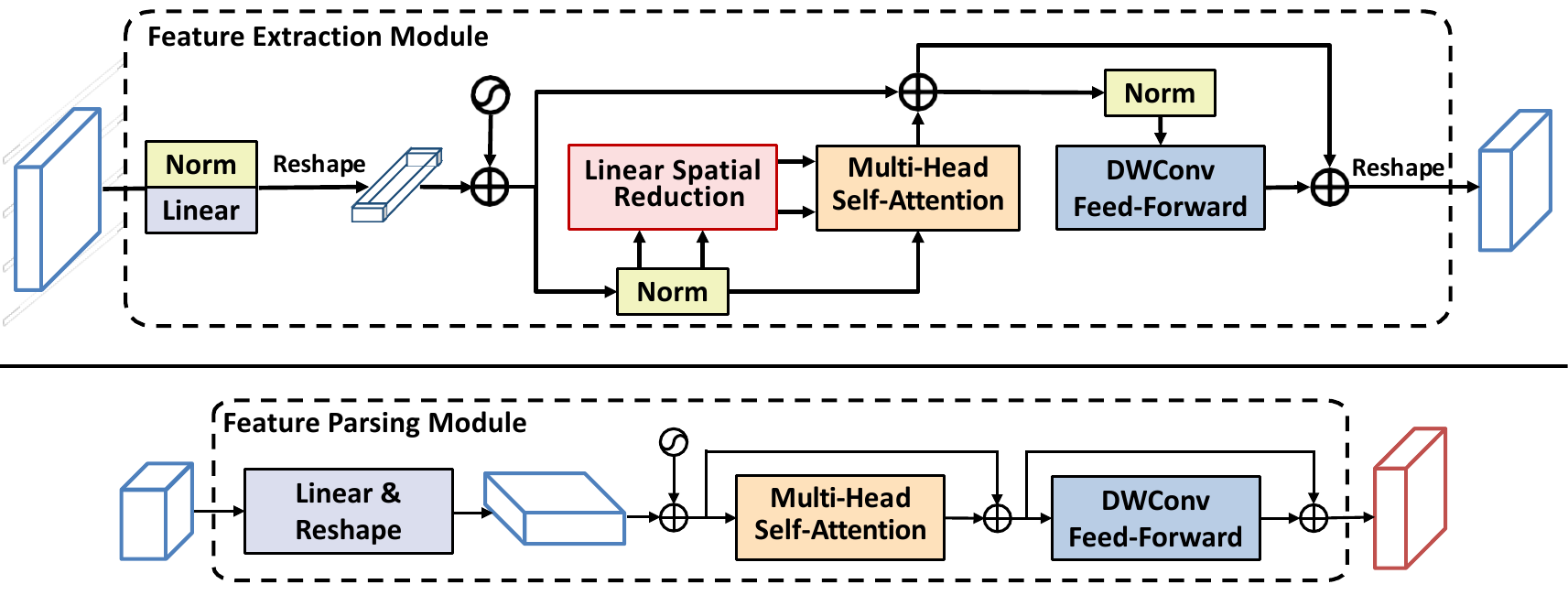}
\caption{\small{The architecture of the Feature Extraction Module (top) in our encoder and Feature Parsing Module (bottom) in our decoder. Zoom in for better visualization.}}
\label{fig:FEM_FPM}
\end{figure*}

In this section, we show the detailed architecture of the Feature Extraction Module in our encoder and the Feature Parsing Module in our decoder in Figure~\ref{fig:FEM_FPM}.


The hyperparameters of backbones in our models are listed as follows:
\begin{itemize}
    \item $S_i$: stride of overlapping patch embedding in Stage $i$;
    \item $C_i$: channel number of output of Stage $i$;
    \item $L_i$: number of encoder layers in Stage $i$;
    \item $R_i$: reduction ratio of SRA layer in Stage $i$;
    \item $P_i$: patch size of Stage $i$;
    \item $N_i$: head number of Efficient Self-Attention in Stage $i$;
    \item $E_i$: expansion ratio of Feed-Forward layer~\citep{vaswani2017attention} in Stage $i$;
\end{itemize}

In addition, we describe a series of PVTv1~\citep{pvt2021} backbones with different scales (Tiny, Small, Medium, and Large) in Table~\ref{tab:arch_pvtv1} and a series of PVTv2~\citep{wang2021pvtv2} backbones with different scales (B1 to B5) in Table~\ref{tab:arch_pvtv2}.

\begin{table*}[!htb]
\centering
\caption{Detailed settings of PVTv1 series which is adopted from~\citep{pvt2021}.}
\renewcommand\arraystretch{1.03}
\setlength{\tabcolsep}{1.0mm}
\resizebox{0.9\textwidth}{!}{%
\begin{tabular}{c|c|c|c|c|c|c}
\toprule
& \textbf{Output Size} & \textbf{Layer Name} & \textbf{PVT-Tiny} & \textbf{PVT-Small} & \textbf{PVT-Medium}  & \textbf{PVT-Large} \\
\midrule \midrule
\multirow{2}{*}[-2.5ex]{\textbf{Stage 1}} & \multirow{2}{*}[-2.5ex]{\scalebox{1.3}{$\frac{H}{4}\times \frac{W}{4}$}} & Patch Embedding & \multicolumn{4}{c}{$P_1=4$;\ \ \ $C_1=64$} \\
\cline{3-7}
& & \tabincell{c}{Transformer\\Encoder} & 
$\begin{bmatrix}
    \begin{array}{l}
    	R_1=8 \\
    	N_1=1 \\
    	E_1=8 \\
    \end{array}
\end{bmatrix} \times 2$ &
$\begin{bmatrix}
	\begin{array}{l}
    	R_1=8 \\
    	N_1=1 \\
    	E_1=8 \\
	\end{array}
\end{bmatrix} \times 3$ &
$\begin{bmatrix}
	\begin{array}{l}
    	R_1=8 \\
    	N_1=1 \\
    	E_1=8 \\
	\end{array}
\end{bmatrix} \times 3$ &
$\begin{bmatrix}
	\begin{array}{l}
    	R_1=8 \\
    	N_1=1 \\
    	E_1=8 \\
	\end{array}
\end{bmatrix} \times 3$ \\
\midrule
\multirow{2}{*}[-2.5ex]{\textbf{Stage 2}} & \multirow{2}{*}[-2.5ex]{\scalebox{1.3}{$\frac{H}{8}\times \frac{W}{8}$}} & Patch Embedding & \multicolumn{4}{c}{$P_2=2$;\ \ \  $C_2=128$} \\
\cline{3-7}
& & \tabincell{c}{Transformer\\Encoder} &
$\begin{bmatrix}
	\begin{array}{l}
    	R_2=4 \\
    	N_2=2 \\
    	E_2=8 \\
	\end{array}
\end{bmatrix} \times 2$ &
$\begin{bmatrix}
	\begin{array}{l}
    	R_2=4 \\
    	N_2=2 \\
    	E_2=8 \\
	\end{array}
\end{bmatrix} \times 3$ &
$\begin{bmatrix}
	\begin{array}{l}
    	R_2=4 \\
    	N_2=2 \\
    	E_2=8 \\
	\end{array}
\end{bmatrix} \times 3$ &
$\begin{bmatrix}
	\begin{array}{l}
    	R_2=4 \\
    	N_2=2 \\
    	E_2=8 \\
	\end{array}
\end{bmatrix} \times 8$\\
\midrule
\multirow{2}{*}[-2.5ex]{\textbf{Stage 3}} & \multirow{2}{*}[-2.5ex]{\scalebox{1.3}{$\frac{H}{16}\times \frac{W}{16}$}} & Patch Embedding & \multicolumn{4}{c}{$P_3=2$;\ \ \  $C_3=320$} \\
\cline{3-7}
& & \tabincell{c}{Transformer\\Encoder} &
$\begin{bmatrix}
	\begin{array}{l}
    	R_3=2 \\
    	N_3=5 \\
    	E_3=4 \\
	\end{array}
\end{bmatrix} \times 2$ &
$\begin{bmatrix}
	\begin{array}{l}
    	R_3=2 \\
    	N_3=5 \\
    	E_3=4 \\
	\end{array}
\end{bmatrix} \times 6$ &
$\begin{bmatrix}
	\begin{array}{l}
    	R_3=2 \\
    	N_3=5 \\
    	E_3=4 \\
	\end{array}
\end{bmatrix} \times 18$ &
$\begin{bmatrix}
	\begin{array}{l}
    	R_3=2 \\
    	N_3=5 \\
    	E_3=4 \\
	\end{array}
\end{bmatrix} \times 27$\\
\midrule
\multirow{2}{*}[-2.5ex]{\textbf{Stage 4}} &  \multirow{2}{*}[-2.5ex]{\scalebox{1.3}{$\frac{H}{32}\times \frac{W}{32}$}} & Patch Embedding & \multicolumn{4}{c}{$P_4=2$;\ \ \ $C_4\!=\!512$} \\
\cline{3-7}
& & \tabincell{c}{Transformer\\Encoder} & 
$\begin{bmatrix}
	\begin{array}{l}
    	R_4=1 \\
    	N_4=8 \\
    	E_4=4 \\
	\end{array}
\end{bmatrix} \times 2$ &
$\begin{bmatrix}
	\begin{array}{l}
    	R_4=1 \\
    	N_4=8 \\
    	E_4=4 \\
	\end{array}
\end{bmatrix} \times 3$ & $\begin{bmatrix}
\begin{array}{l}
    	R_4=1 \\
    	N_4=8 \\
    	E_4=4 \\
	\end{array}
\end{bmatrix} \times 3$ & $\begin{bmatrix}
\begin{array}{l}
    	R_4=1 \\
    	N_4=8 \\
    	E_4=4 \\
	\end{array}
\end{bmatrix} \times 3$\\
\bottomrule
\end{tabular}
}
\label{tab:arch_pvtv1}
\end{table*}

\begin{table*}[t]
\centering
\caption{Detailed settings of PVTv2 series which is adopted from~\citep{wang2021pvtv2}.}
\renewcommand\arraystretch{1.0}
\setlength{\tabcolsep}{3.0mm}
\resizebox{0.9\textwidth}{!}{%
\begin{tabular}{c|c|c|H c|c|H c|c|c}
\toprule
& \textbf{Output Size} & \textbf{Layer Name} & \textbf{PVT-B0} & \textbf{PVT-B1} & \textbf{PVT-B2} & \textbf{PVT-B2-Li} & \textbf{PVT-B3} & \textbf{PVT-B4} & \textbf{PVT-B5}\\
\midrule \midrule
\multirow{3}{*}[-3.5ex]{\textbf{Stage 1}} & \multirow{3}{*}[-3.5ex]{\scalebox{1.3}{$\frac{H}{4}\times \frac{W}{4}$}} & \multirow{2}{*}{\tabincell{c}{Overlapping\\Patch Embedding}} &\multicolumn{7}{c}{$S_1=4$} \\
\cline{4-10}
& & & $C_1=32$ & \multicolumn{6}{c}{$C_1=64$}\\
\cline{3-10}
& & \tabincell{c}{Transformer\\Encoder} & 
$\begin{matrix}
	R_1=8 \\
	N_1=1 \\
	E_1=8 \\
	L_1=2 \\
\end{matrix}$ &
$\begin{matrix}
	R_1=8 \\
	N_1=1 \\
	E_1=8 \\
	L_1=2 \\
\end{matrix}$ &
$\begin{matrix}
	R_1=8 \\
	N_1=1 \\
	E_1=8 \\
	L_1=3 \\
\end{matrix}$ &
$\begin{matrix}
	P_1=7 \\
	N_1=1 \\
	E_1=8 \\
	L_1=3 \\
\end{matrix}$ &
$\begin{matrix}
	R_1=8 \\
	N_1=1 \\
	E_1=8 \\
	L_1=3 \\
\end{matrix}$ &
$\begin{matrix}
	R_1=8 \\
	N_1=1 \\
	E_1=8 \\
	L_1=3 \\
\end{matrix}$ &
$\begin{matrix}
	R_1=8 \\
	N_1=1 \\
	E_1=4 \\
	L_1=3 \\
\end{matrix}$\\
\midrule
\multirow{3}{*}[-3.5ex]{\textbf{Stage 2}} & \multirow{3}{*}[-3.5ex]{\scalebox{1.3}{$\frac{H}{8}\times \frac{W}{8}$}} & \multirow{2}{*}{\tabincell{c}{Overlapping\\Patch Embedding}} & \multicolumn{7}{c}{$S_2=2$} \\
\cline{4-10}
& & & $C_2=64$ & \multicolumn{6}{c}{$C_2=128$}\\
\cline{3-10}
& & \tabincell{c}{Transformer\\Encoder} &
$\begin{matrix}
	R_2=4 \\
	N_2=2 \\
	E_2=8 \\
	L_2=2 \\
\end{matrix}$ &
$\begin{matrix}
	R_2=4 \\
	N_2=2 \\
	E_2=8 \\
	L_2=2 \\
\end{matrix}$ &
$\begin{matrix}
	R_2=4 \\
	N_2=2 \\
	E_2=8 \\
	L_2=3 \\
\end{matrix}$ &
$\begin{matrix}
	P_2=7 \\
	N_2=2 \\
	E_2=8 \\
	L_2=3 \\
\end{matrix}$ &
$\begin{matrix}
	R_2=4 \\
	N_2=2 \\
	E_2=8 \\
	L_2=3 \\
\end{matrix}$ &
$\begin{matrix}
	R_2=4 \\
	N_2=2 \\
	E_2=8 \\
	L_2=8 \\
\end{matrix}$ &
$\begin{matrix}
	R_2=4 \\
	N_2=2 \\
	E_2=4 \\
	L_2=6 \\
\end{matrix}$\\
\midrule
\multirow{3}{*}[-3.5ex]{\textbf{Stage 3}} & \multirow{3}{*}[-3.5ex]{\scalebox{1.3}{$\frac{H}{16}\times \frac{W}{16}$}} & \multirow{2}{*}{\tabincell{c}{Overlapping\\Patch Embedding}} & \multicolumn{7}{c}{$S_3=2$} \\
\cline{4-10}
& & & $C_3=160$ & \multicolumn{6}{c}{$C_3=320$}\\
\cline{3-10}
& & \tabincell{c}{Transformer\\Encoder} &
$\begin{matrix}
	R_3=2 \\
	N_3=5 \\
	E_3=4 \\
	L_3=2 \\
\end{matrix}$ &
$\begin{matrix}
	R_3=2 \\
	N_3=5 \\
	E_3=4 \\
	L_3=2 \\
\end{matrix}$ &
$\begin{matrix}
	R_3=2 \\
	N_3=5 \\
	E_3=4 \\
	L_3=6 \\
\end{matrix}$ &
$\begin{matrix}
	P_3=7 \\
	N_3=5 \\
	E_3=4 \\
	L_3=6 \\
\end{matrix}$ &
$\begin{matrix}
	R_3=2 \\
	N_3=5 \\
	E_3=4 \\
	L_3=18 \\
\end{matrix}$ &
$\begin{matrix}
	R_3=2 \\
	N_3=5 \\
	E_3=4 \\
	L_3=27 \\
\end{matrix}$ &
$\begin{matrix}
	R_3=2 \\
	N_3=5 \\
	E_3=4 \\
	L_3=40 \\
\end{matrix}$\\
\midrule
\multirow{3}{*}[-3.5ex]{\textbf{Stage 4}} &  \multirow{3}{*}[-3.5ex]{\scalebox{1.3}{$\frac{H}{32}\times \frac{W}{32}$}} & \multirow{2}{*}{\tabincell{c}{Overlapping\\Patch Embedding}} & \multicolumn{7}{c}{$S_4=2$} \\
\cline{4-10}
& & & $C_4=256$ & \multicolumn{6}{c}{$C_4=512$}\\
\cline{3-10}
& & \tabincell{c}{Transformer\\Encoder} & 
$\begin{matrix}
	R_4=1 \\
	N_4=8 \\
	E_4=4 \\
	L_4=2 \\
\end{matrix}$ &
$\begin{matrix}
	R_4=1 \\
	N_4=8 \\
	E_4=4 \\
	L_4=2 \\
\end{matrix}$ &
$\begin{matrix}
	R_4=1 \\
	N_4=8 \\
	E_4=4 \\
	L_4=3 \\
\end{matrix}$ & 
$\begin{matrix}
	P_4=7 \\
	N_4=8 \\
	E_4=4 \\
	L_4=3 \\
\end{matrix}$ & 
$\begin{matrix}
	R_4=1 \\
	N_4=8 \\
	E_4=4 \\
	L_4=3 \\
\end{matrix}$ & 
$\begin{matrix}
	R_4=1 \\
	N_4=8 \\
	E_4=4 \\
	L_4=3 \\
\end{matrix}$ &
$\begin{matrix}
	R_4=1 \\
	N_4=8 \\
	E_4=4 \\
	L_4=3 \\
\end{matrix}$\\
\bottomrule
\end{tabular}
}
\label{tab:arch_pvtv2}
\end{table*}


\section{Experimental Setups}
\label{app-sec:setup}

\subsection{Datasets}
We comprehensively evaluated our proposed method on diverse datasets to demonstrate its exceptional performance and versatility. These datasets encompass a broad spectrum of segmentation tasks such as Glass (Transparent) datasets (Trans10k-v2~\cite{xie2021segmenting}, RGBP-Glass~\cite{PGSNet2022}, and GSD-S~\cite{gsds2022}), Mirror (Reflection) datasets (MSD~\cite{msd2019}, PMD~\cite{pmd2020}, and RGBD-Mirror~\cite{rgbdm2021}), and generic datasets, which consists of both glass and mirror objects (TROSD~\cite{trosd2023}, and Stanford2D3D~\cite{s2d3d}), ranging from binary to semantic segmentation, with a particular focus on images featuring reflective, transparent, or both characteristics. Our evaluation also considers the varied positions and fields of view (FOV) of objects within the images. Objects of interest may appear near or far from the camera's perspective, positioned randomly or at the center of the frame, providing a rich and realistic testing environment. Furthermore, the datasets we utilized are substantial in size, ensuring coverage of a broad range of environmental and scenario complexities. This encompasses indoor and outdoor scenarios, varying lighting conditions, diverse object scales, different viewpoints, and levels of occlusion. Our extensive evaluation showcases the robustness and adaptability of our method across a wide array of real-world conditions. Details of each dataset are shown in Table~\ref{tab:datasets}.

\begin{table*}[!t]
\centering
\setlength{\tabcolsep}{6pt}
\renewcommand{\arraystretch}{1.4}
\caption{Comparison between different datasets in our experiments. ``P'', ``D'', and ``S'' denote polarization images, depth images, and semantic maps, respectively. Note that our method \textbf{\underline{uses only RGB images}} as input for both training and testing.}
\label{tab:datasets}
\resizebox{0.8\textwidth}{!}{%
\begin{tabular}{@{}c|l|l|c|c|c|c|c@{}}
\toprule
\textbf{} & \textbf{Dataset} & \textbf{Modalities} & \textbf{No. of Images} & \textbf{Tasks} & \textbf{Types} & \textbf{FOV} & \textbf{Position} \\ 
\midrule \midrule
\multirow{3}{*}{Glass} & Trans10k-v2~\cite{xie2021segmenting}      & RGB                 & 10,428          & semantic       & both    & both         & random            \\
& RGBP-Glass~\cite{PGSNet2022}       & RGB-P               & 4,511           & binary         & transparent    & far          & random            \\
& GSD-S~\cite{gsds2022}            & RGB-S               & 4,519 
& binary         & transparent    & far          & random            \\ \midrule
\multirow{3}{*}{Mirror} & MSD~\cite{msd2019}              & RGB                 & 4,018           & binary         & reflective     & both         & center            \\
& PMD~\cite{pmd2020}              & RGB                 & 6,461           & binary         & reflective     & both         & random            \\
& RGBD-Mirror~\cite{rgbdm2021}      & RGB-D               & 3,049           & binary         & reflective     & both         & center            \\ \midrule
\multirow{2}{*}{Generic} & TROSD~\cite{trosd2023}            & RGB-D               & 11,060          & semantic       & both           & near         & center            \\
& Stanford2D3D~\cite{s2d3d}     & RGB-D               & 70,496          & semantic       & both           & far          & random            \\ \bottomrule
\end{tabular}%
}
\end{table*}

\subsection{Implementation Details}
We implemented our method in PyTorch 1.8.0 and CUDA 11.2. We adopted AdamW optimizer~\cite{adamw} where the learning rate $\gamma$ was set to $10^{-4}$ with epsilon $10^{-8}$ and weight decay $10^{-4}$. Our model was trained with a batch size of 8 and on a single NVIDIA RTX 3090 GPU, but it can still be trained on an older 2080 Ti or 1080 Ti GPU with a smaller batch size, e.g., 4. We evaluated all variants of our network on the validation set at every epoch during training. We used the best model of each variant on the validation set to evaluate the variant on the test set. The training process was completed once no further improvements were achieved. We use mean Intersection over Union (mIoU) as the primary evaluation metric to assess segmentation performance.

\subsection{Evaluation metrics.} 
We adopt four widely used metrics from~\cite{PGSNet2022} to assess glass segmentation performance quantitatively: mean intersection over union (mIoU), weighted F-measure ($F_\beta^w$), mean absolute error (MAE), and balance error rate (BER).

\noindent\textbf{Intersection over Union} $(IoU)$ is a widely used metric in segmentation tasks, which is defined as:
\vspace{-1mm}
\begin{equation}
    IoU=\frac{\sum\limits_{i=1}^H \sum\limits_{j=1}^W(G(i, j) * P_b(i, j))}{\sum\limits_{i=1}^H \sum\limits_{j=1}^W(G(i, j)+P_b(i, j)-G(i, j) * P_b(i, j))}
\end{equation}
where $G$ is the ground truth mask with values of the glass region being one while those of the non-glass region are 0; $P_b$ is the predicted mask binarized with a threshold of 0.5; and $H$ and $W$ are the height and width of the ground truth mask, respectively.

\noindent\textbf{Weighted F-measure} $(F_\beta^w)$ is adopted from the salient object detection tasks with $\beta = 0.3$. F-measure $(F_\beta)$ is a measure of both the precision and recall of the prediction map. Recent studies~\cite{Fan_2017_ICCV}  have suggested that the weighted F-measure $(F_\beta^w)$~\cite{Margolin_2014_CVPR} can provide more reliable evaluation results than the traditional $F_\beta$. Thus, we report $F_\beta^w$ in the comparison.

\noindent\textbf{Mean Absolute Error} (MAE) is widely used in foreground-background segmentation tasks, which calculates the element-wise difference between the prediction map $P$ and the ground truth mask $G$ :
\vspace{-1mm}
\begin{equation}
    MAE=\frac{1}{H \times W} \sum_{i=1}^H \sum_{j=1}^W|P(i, j)-G(i, j)|,
\end{equation}
where $P(i, j)$ indicates the predicted probability score at location $(i, j)$.

\noindent\textbf{Balance Error Rate} (BER) is a standard metric used in shadow detection tasks, defined as:
\begin{equation}
    BER=(1-\frac{1}{2}(\frac{T P}{N_p}+\frac{T N}{N_n})) \times 100
\end{equation}
where $T P, T N, N_p$, and $N_n$ represent the numbers of true positive, true negative, glass, and non-glass pixels, respectively.

\subsection{Qualitative and Quantitative Results}
We evaluated the performance of our method across three distinct tasks: glass segmentation, mirror segmentation, and generic segmentation. 
To ensure \textbf{fair comparisons}, we have carefully selected our model variants (Ours-X with X is postfixes: -T, -S, -M, -L, -B1, -B2, -B3, -B4, and -B5, represented the size of the model as PVTv1 Tiny, Small, Medium, Large, and PVTv2 B1-5, respectively) that have \textbf{similar model's size or complexity} used by other methods, as indicated in the respective tables. 

\section{Additional Experiments}
\label{app-sec:exp}

\subsection{Comparison on Glass Object Segmentation}

We benchmarked our method against recent glass segmentation methods on the binary (RGBP-Glass and GSD-S dataset) and semantic segmentation (Trans10K-v2 dataset) tasks.

\begin{table}[!t]
\footnotesize
\centering
\caption{\small{Quantitative comparison against SOTAs on RGB-P dataset~\cite{PGSNet2022}. ($\ast$) denotes the glass segmentation methods with additional input polarization images.}}
\setlength{\tabcolsep}{3pt}
\renewcommand{\arraystretch}{1.1}
\resizebox{0.46\textwidth}{!}{%
\begin{tabular}{@{}lH|ccccc@{}}
\toprule
\textbf{Method} & \textbf{Backbone} & \textbf{GFLOPs}~$\downarrow$ & \textbf{mIoU}~$\uparrow$ & \textbf{$F_\beta^w$}~$\uparrow$ & \textbf{MAE}~$\downarrow$ & \textbf{BER}~$\downarrow$ \\ \midrule \midrule
EAFNet~\cite{Xiang2021} $\ast$          & ResNet-18        & 18.93          & 53.86        & 0.611      & 0.237        & 24.65        \\
PM R-CNN~\cite{kalra2020} $\ast$     & ResNet-101        & 56.59          & 66.03        & 0.714      & 0.178        & 18.92        \\ 
PGSNet~\cite{PGSNet2022} $\ast$          & Conformer-B       & \underline{290.62}         & 81.08        & 0.842      & 0.091        & \underline{9.63}         \\ 
\midrule

Trans2Seg~\cite{xie2021segmenting}        & ResNet-50         & 49.03          & 75.21        & 0.799      & 0.122        & 13.23        \\
TransLab~\cite{xie2020segmenting}         & ResNet-50         & 61.26          & 73.59        & 0.772      & 0.148        & 15.73        \\
SegFormer~\cite{Xie2021SegFormerSA}       & MiT-B5            & 70.24           & 78.42        & 0.815      & 0.121        & 13.03        \\
GSD~\cite{Jiaying2021}             & ResNeXt-101       & 92.69          & 78.11        & 0.806      & 0.122        & 12.61        \\
\rowcolor{gray!15} Ours-B5           & PVTv2-B5       & 154.37         & \underline{82.77}        & \underline{0.879}      & \textbf{0.042}        & \textbf{9.59}         \\ 
GDNet~\cite{Mei_2020_CVPR}           & ResNeXt-101       & 271.53         & 77.64        & 0.807      & 0.119        & 11.79        \\
SETR~\cite{setr}             & ViT-Large         & 240.11          & 77.60         & 0.817      & 0.114        & 11.46        \\
PanoGlassNet~\cite{chang2024panoglassnet} & - & \textbf{581.04} & \textbf{86.89} & \textbf{0.929} & \underline{0.068} & - \\
\bottomrule
\end{tabular}%
}
\vspace{-2mm}
\label{tab:rgbp_full}
\end{table}

\begin{figure*}[!t]
    \centering
    \includegraphics[width=0.99\linewidth]{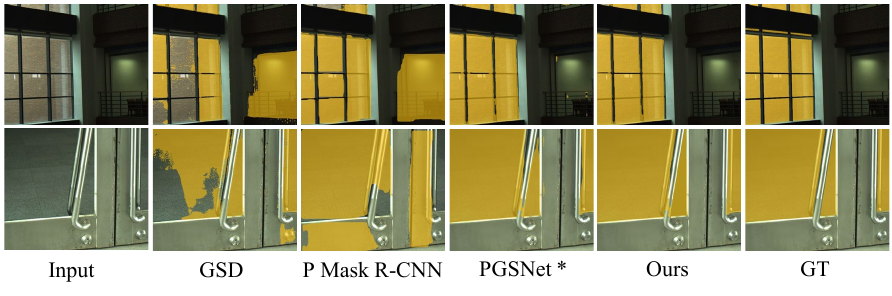}
    \caption{\small{Qualitative comparison of our method with other methods on RGB-P dataset~\cite{PGSNet2022}. ($\ast$) denotes the glass segmentation method with additional polarization images as input.}}
    \label{fig:rgbp}
\end{figure*}

\vspace{1mm}
\noindent\textbf{RGBP-Glass dataset.}
We extensively compare the effectiveness of our method with state-of-the-art methods, as shown in Table~\ref{tab:rgbp_full}. All methods are retrained on the RGBP-Glass dataset~\cite{PGSNet2022} for a fair comparison. EAFNet~\cite{Xiang2021}, Polarized Mask R-CNN (P.M. R-CNN)~\cite{kalra2020}, and PGSNet~\cite{PGSNet2022} are the three methods that leverage polarization cues. SETR~\cite{setr}, SegFormer~\cite{Xie2021SegFormerSA} are the two methods focusing on general semantic/instance segmentation tasks. GDNet~\cite{Mei_2020_CVPR}, TransLab~\cite{xie2020segmenting}, Trans2Seg~\cite{xie2021segmenting}, and GSD~\cite{Jiaying2021} and our method are in-the-wild glass segmentation methods but only rely on RGB input. From Figure~\ref{fig:rgbp}, we can see that our method outperforms all other methods. It should be noted that our method outperforms previous works that utilize additional input signals, such as polarization cues~\cite{Xiang2021,kalra2020,PGSNet2022}, while remaining efficient.

\begin{figure*}[!t]
    \centering
    \includegraphics[width=0.99\linewidth]{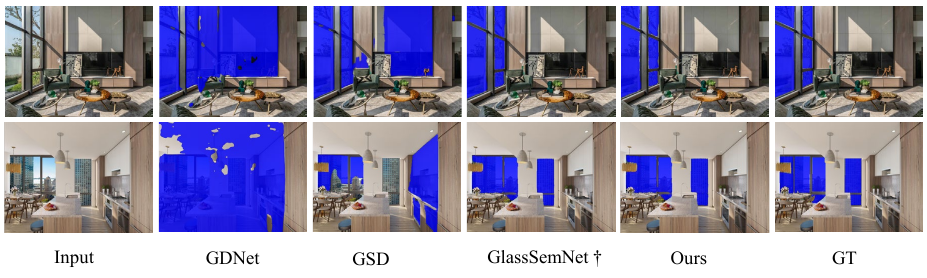}
    \caption{\small{Qualitative comparison of our method with other methods on GSD-S dataset~\cite{gsds2022}. ($\dagger$) denotes the glass segmentation method with semantic context and post-processing refinement.}}
    \label{fig:gsds}
\end{figure*}

\vspace{1mm}
\noindent\textbf{GSD-S dataset.}
We compare our method with other recent methods in Table~\ref{tab:gsds_full} and Figure~\ref{fig:gsds}, includes generic semantic segmentation methods (PSPNet~\cite{pspnet}, DeepLabV3+~\cite{Chen2018eccv}, PSANet~\cite{zhao2018psanet}, DANet~\cite{danet}), recent state-of-the-art models that utilize transformer technique (SETR~\cite{setr}, Swin~\cite{swin}, SegFormer~\cite{Xie2021SegFormerSA}, Twins~\cite{twins2021}), and glass surface detection methods (GDNet~\cite{Mei_2020_CVPR}, GSD~\cite{Jiaying2021}, GlassSemNet~\cite{gsds2022}). For a fair comparison, all methods are retrained on the GSD-S dataset~\cite{gsds2022}. Our method outperforms all other methods and achieves comparable performance to GlassSemNet~\cite{gsds2022}, which provides additional semantic context information. GlassSemNet~\cite{gsds2022} points out that humans frequently use the semantic context of their surroundings to reason, as this provides information about the types of things to be found and how close they might be to one another. For instance, glass windows are more likely to be found close to other semantically related objects (walls and curtains) than to things (cars and trees). Their method uses semantic context information as an additional input to progressively learn contextual correlations among objects, both spatially and semantically, thereby boosting performance. Their predictions are then refined by Fully Connected Conditional Random Fields~\cite{crf_2021} to improve performance further.

\begin{table}[!t]
\footnotesize
\centering
\caption{\small{Evaluation results on GSD-S dataset~\cite{gsds2022}. \textbf{Note that:} we use only RGB as input to our method. ($\dagger$) denotes the glass segmentation method with additional semantic context information and post-processing refinement.}}
\vspace{-1mm}
\setlength{\tabcolsep}{6pt}
\renewcommand{\arraystretch}{1.1}
\resizebox{0.46\textwidth}{!}{%
\begin{tabular}{l|cccc}
\toprule
\textbf{Method} & \textbf{mIoU}~$\uparrow$ & \textbf{$F_\beta^w$}~$\uparrow$ & \textbf{MAE}~$\downarrow$ & \textbf{BER}~$\downarrow$ \\ 
\midrule \midrule
PSPNet~\cite{pspnet}           & 56.1        & 0.679      & 0.093        & 13.41         \\
DeepLabV3+~\cite{Chen2018eccv}       & 55.7        & 0.671      & 0.100        & 13.11        \\
PSANet~\cite{zhao2018psanet}           & 55.1        & 0.656      & 0.104        & 12.61        \\
DANet~\cite{danet}            & 54.3        & 0.673      & 0.098        & 14.78        \\
SCA-SOD~\cite{SCA_SOD}         & 55.8        & 0.689      & 0.087        & 15.03        \\
\midrule
SETR~\cite{setr}             & 56.7        & 0.679      & 0.086        & 13.25        \\
Segmenter~\cite{segmenter}        & 53.6        & 0.645      & 0.101        & 14.02        \\
Swin~\cite{swin}             & 59.6        & 0.702      & 0.082        & 11.34        \\
T-2-T ViT~\cite{T2T_ViT}              & 56.2        & 0.693      & 0.087        & 14.72        \\
SegFormer~\cite{Xie2021SegFormerSA}        & 54.7        & 0.683      & 0.094        & 15.15        \\
Twins~\cite{twins2021}            & 59.1        & 0.703      & 0.084        & 12.43        \\
\midrule
GDNet~\cite{Mei_2020_CVPR}           & 52.9        & 0.642      & 0.101        & 18.17        \\
GSD~\cite{Jiaying2021}         & 72.1        & 0.821      & 0.061        & 10.02        \\
VBNet~\cite{qi2024glass}      & 73.5       & 0.837      & \underline{0.038}       & 10.07 \\
\rowcolor{gray!15} Ours-B5                          & \underline{75.2}        & \underline{0.859}       & 0.046        & \textbf{9.04}         \\ 
GlassSemNet~\cite{gsds2022} $\mathbf{\dagger}$      & \textbf{75.3}        & \textbf{0.860}       & \textbf{0.035}        & \underline{9.26}         \\ 
\bottomrule
\end{tabular}%
}
\label{tab:gsds_full}
\vspace{-3mm}
\end{table}

\begin{table}[!t]
\footnotesize
\centering
\caption{\small{Quantitative evaluation of our method and existing methods on the Trans10K-v2 dataset~\cite{xie2021segmenting}. 
}}
\setlength{\tabcolsep}{4pt}
\renewcommand{\arraystretch}{1.1}
\resizebox{0.48\textwidth}{!}{%
\begin{tabular}{@{}lcccc@{}}
\toprule
\textbf{Method} & \textbf{GFLOPs}~$\downarrow$ & \textbf{MParams}~$\downarrow$ & \textbf{ACC}~$\uparrow$ & \textbf{mIoU}~$\uparrow$ \\ 
\midrule \midrule
HRNet\_w18~\cite{hrnet}  & 4.20 & 1.53 & 89.58 & 54.25 \\
LEDNet~\cite{lednet}  & 6.23 & - & 86.07 & 46.40 \\
Trans4Trans-T~\cite{zhang2022trans4trans} &  10.45 & - &  \underline{93.23}  & \underline{68.63} \\
\rowcolor{gray!15} Ours-T &  10.50 & 12.72 &  \textbf{93.52}  & \textbf{69.53} \\
ICNet~\cite{icnet}  & 10.64 & 8.46 & 78.23 & 23.39 \\
\midrule
BiSeNet~\cite{bisenet}  & 19.91 & 13.3 & 89.13 & 58.40 \\
Trans4Trans-S~\cite{zhang2022trans4trans} &  19.92 & - &  94.57  & 74.15 \\ 
\rowcolor{gray!15} Ours-S &  20.00 & 23.98 &  \underline{94.83}  & \underline{75.32} \\ 
\rowcolor{gray!15} Ours-B1 &  21.29 & 14.87 &  \textbf{95.37}  & \textbf{77.05} \\ 
\midrule
Trans4Trans-M~\cite{zhang2022trans4trans} &  34.38 & - &  95.01  & 75.14 \\ 
\rowcolor{gray!15} Ours-M &  34.51 & 43.70 &  95.08  & 76.06 \\
DenseASPP~\cite{denseaspp} & 36.20 & 29.09 & 90.86 & 63.01 \\
\rowcolor{gray!15} Ours-B2 &  37.03 & 27.59 &  \textbf{95.92}  & \textbf{79.29} \\ 
DeepLabv3+~\cite{Chen2018eccv} & 37.98 & 28.74 & 92.75 & 68.87 \\
FCN~\cite{fcn2015}  & 42.23 & 34.99 & 91.65 & 62.75 \\
RefineNet~\cite{refinenet}  & 44.56 & 29.36 & 87.99 & 58.18 \\
Trans2Seg~\cite{xie2021segmenting} &  49.03 & 56.20 &  94.14  & 72.15 \\ 
\rowcolor{gray!15} Ours-L &  50.54 & 60.86 &  \underline{95.28}  & \underline{77.35} \\
\midrule
TransLab~\cite{xie2020segmenting}  & 61.31 & 42.19 & 92.67  & 69.00 \\  
\rowcolor{gray!15} Ours-B3 &  68.35 & 51.21 &  \underline{96.28}  & \underline{80.04} \\ 
\rowcolor{gray!15} Ours-B4 &  79.34 & 67.11 &  \textbf{96.59}  & \textbf{80.99} \\ 
\midrule
To-Former-B2~\cite{chen2024toformer}  & 117.74 & - & - & 77.43 \\
U-Net~\cite{unet}  & 124.55 & 13.39 & 81.90 & 29.23 \\
DUNet~\cite{dunet}  & 123.69 & - & 90.67 & 59.01 \\
\rowcolor{gray!15} Ours-B5 &  154.37 & 106.19 &  \textbf{96.93}  & \textbf{81.37} \\ 
DANet~\cite{danet}  & 198.00 & - & \underline{92.70}  & \underline{68.81} \\
PSPNet~\cite{pspnet}  & 187.03 & 50.99 & 92.47 & 68.23 \\
\bottomrule
\end{tabular}
}
\label{tab:trans10k_sota_full}
\vspace{-2mm}
\end{table}

\begin{figure}[!t]
    \centering
    \includegraphics[width=0.96\linewidth]{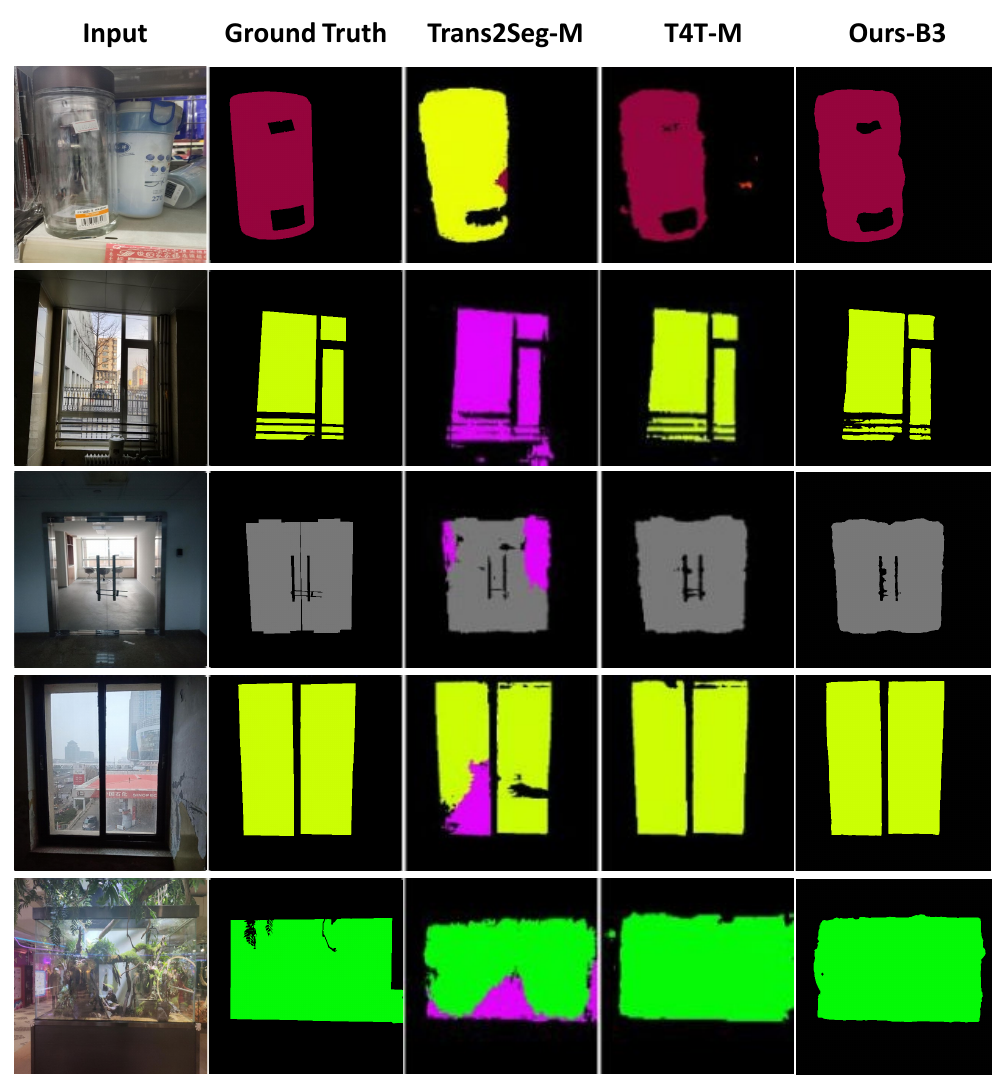}
    \caption{\small{Qualitative comparison of our method and existing methods on Trans10K-v2~\cite{xie2021segmenting}. For a fair comparison, we used Ours-B3, which has the same network size as other methods (-M model).}}
    \label{fig:vis_tran10k_full}
    \vspace{-2mm}
\end{figure}

\vspace{1mm}
\noindent\textbf{Trans10k-v2 dataset.}
Shifting our focus to the semantic glass segmentation task, where the challenge extends beyond merely detecting glass areas to classifying them into 11 fine-grained categories, our method still reigns supreme, as shown in Table~\ref{tab:trans10k_sota_full}. Figure~\ref{fig:vis_tran10k_full} also confirms that 
our method achieves higher segmentation quality with better transparent features, e.g., the segmentation of two overlapping doors is accurately obtained.
These comprehensive evaluations underscore the effectiveness of our approach across diverse glass segmentation scenarios, affirming its position as a top-performing and computationally efficient choice for these tasks.

\subsubsection{Comparison on Binary Mirror Segmentation}

\vspace{1mm}
\noindent\textbf{MSD and PMD datasets.}
We compare quantitative results of the state-of-the-art methods and our method on MSD and PMD datasets, including four RGB salient object detection methods CPDNet~\cite{wu2019cascaded}, MINet~\cite{pang2020multi}, LDF~\cite{Wei_2020_CVPR}, and VST~\cite{Liu_2021_ICCV}, and five mirror detection methods MirrorNet~\cite{msd2019}, PMDNet~\cite{pmd2020}, SANet~\cite{SANet2022}, VCNet~\cite{VCNet2023}, SATNet~\cite{SATNet2023}. As shown in Table~\ref{tab:msd&pmd}, our method achieves the best performance in terms of all the evaluation metrics. Significantly, we outperform the second-best method by 5.63\% on the MSD dataset.

\vspace{1mm}
\noindent\textbf{RGBD-Mirror dataset.}
Our method is also compared with seven RGB-D salient object detection methods such as HDFNet~\cite{pang2020hierarchical}, S2MA~\cite{Liu_2020_CVPR}, JL-DCF~\cite{Fu_2020_CVPR}, DANet~\cite{danet}, BBSNet~\cite{fan2020bbs} and VST~\cite{Liu_2021_ICCV}, and four mirror detection methods, including PDNet~\cite{rgbdm2021}, SANet~\cite{SANet2022}, VCNet~\cite{VCNet2023}, and PDNet~\cite{rgbdm2021} on the RGBD-Mirror dataset. Our method outperforms all competing methods, even though we do not use depth information, as shown in Table~\ref{tab:rgbdm}. 

\begin{table}[!t]
\centering
\footnotesize
\caption{\small{Quantitative results of our method with SOTAs on Salient Object Detection (the first five methods) and Mirror Detection (the last ten methods) on MSD and PMD datasets. 
}}
\vspace{-1mm}
\setlength{\tabcolsep}{4pt}
\renewcommand{\arraystretch}{1.1}
\resizebox{0.48\textwidth}{!}{%
\begin{tabular}{l|ccc|ccc}
\toprule
\multirow{2}{*}{\textbf{Method}} & \multicolumn{3}{c|}{\textbf{MSD}}   & \multicolumn{3}{c}{\textbf{PMD}} \\ 
\cmidrule(l){2-7} 
                                 & \textbf{IoU}~$\uparrow$ & \textbf{$F_\beta^w$}~$\uparrow$ & \textbf{MAE}~$\downarrow$                & \textbf{IoU}~$\uparrow$ & \textbf{$F_\beta^w$}~$\uparrow$ & \textbf{MAE}~$\downarrow$                \\ 
\midrule \midrule
CPDNet~\cite{wu2019cascaded}                           & 57.58                          & 0.743                          & 0.115                          & 60.04                          & 0.733                          & 0.041                          \\
MINet~\cite{pang2020multi}                            & 66.39                          & 0.823                          & 0.087                          & 60.83                          & 0.798                          & 0.037                          \\
LDF~\cite{Wei_2020_CVPR}        & 72.88                          & 0.843                          & 0.068                          & 63.31                          & 0.796                          & 0.037                          \\
VST~\cite{Liu_2021_ICCV}        & 79.09                          & 0.867                          & 0.052                          & 59.06                          & 0.769                          & 0.035                          \\ 
ShadowSAM~\cite{wang2024shadow}        & -      & 0.700          & 0.080        & -         & 0.685        & 0.095       \\ 
\midrule
MirrorNet~\cite{msd2019}   & 78.88                          & 0.856                          & 0.066                          & 58.51                          & 0.741                          & 0.043                          \\
PMDNet~\cite{pmd2020}                           & 81.54                          & 0.892                          & 0.047                          & 66.05                          & 0.792                          & 0.032                          \\
SANet~\cite{SANet2022}                            & 79.85                          & 0.879                          & 0.054                          & 66.84                          & 0.837                          & 0.032                          \\
HetNet~\cite{HetNet2023}                            & 82.80                          & 0.906                          & 0.043                          & 69.00                          & 0.814                          & 0.029                          \\
UTLNeT~\cite{zhou2023utlnet}                            & 83.05                          & 0.892                          & 0.040                          & -                          & -                          & -                          \\
WSMD~\cite{zha2024weakly}                            & 75.00                          & 0.780                          & 0.078                          & 60.00                          & 0.630                          & 0.051                          \\
DPRNet~\cite{zha2024dual}                            & \underline{86.60}                          & 0.888                          & 0.033                          & \textbf{72.10}                          & 0.766                          & 0.026                          \\
VCNet~\cite{VCNet2023}                            & 80.08                          & 0.898                          & 0.044                          & 64.02                          & 0.815                          & 0.028                          \\
SATNet~\cite{SATNet2023}                           & 85.41                          & \underline{0.922}                          & 0.033                          & \underline{69.38}                          & \underline{0.847}                          & \underline{0.025}                          \\ 
CSFwinformer~\cite{xie2024csf}        & -      & 0.865        & \underline{0.030}         & -           & 0.836        & 0.039       \\ 
\midrule
\rowcolor{gray!15} Ours-B3                             & \textbf{91.04} & \textbf{0.953} & \textbf{0.028} & \underline{69.61} & \textbf{0.853} & \textbf{0.021} \\ \bottomrule
\end{tabular}%
}
\vspace{-3mm}
\label{tab:msd&pmd}
\end{table}

\begin{table}[!t]
\centering
\footnotesize
\caption{\small{Quantitative results of SOTAs on RGBD-Mirror dataset.}}
\setlength{\tabcolsep}{7pt}
\renewcommand{\arraystretch}{1.1}
\resizebox{0.44\textwidth}{!}{%
\begin{tabular}{l|c|ccc}
\toprule
\textbf{Method} & \textbf{Input} & \textbf{IoU}~$\uparrow$ & $\mathbf{F_{\beta}^w}$~$\uparrow$ & \textbf{MAE}~$\downarrow$ \\ \midrule \midrule
HDFNet~\cite{pang2020hierarchical}          & RGB-D     & 44.73         & 0.733             & 0.093           \\
S2MA~\cite{Liu_2020_CVPR}            & RGB-D     & 60.87         & 0.781             & 0.070           \\
DANet~\cite{danet}           & RGB-D     & 67.81         & 0.835             & 0.060           \\
JL-DCF~\cite{Fu_2020_CVPR}          & RGB-D     & 69.65         & 0.844             & 0.056           \\
VST~\cite{Liu_2021_ICCV}             & RGB-D     & 70.20         & 0.851             & 0.052           \\
BBSNet~\cite{fan2020bbs}          & RGB-D     & 74.33         & 0.868             & 0.046           \\
PDNet~\cite{rgbdm2021}           & RGB-D     & 77.77         & 0.878             & 0.041           \\
UTLNet~\cite{zhou2023utlnet}         & RGB-D                   & \underline{80.50}                          & 0.858                          & 0.032 \\
\midrule
VCNet~\cite{VCNet2023}           & RGB       & 73.01         & 0.849             & 0.052           \\
PDNet~\cite{rgbdm2021}           & RGB       & 73.57         & 0.851             & 0.053           \\
SANet~\cite{SANet2022}           & RGB       & 74.99         & 0.873             & 0.048           \\
SATNet~\cite{SATNet2023}          & RGB       & 78.42         & \underline{0.906}             & \underline{0.031}           \\
WSMD~\cite{zha2024weakly}         & RGB                   & 61.60                          & 0.655                          & 0.088 \\
DPRNet~\cite{zha2024dual}         & RGB                   & 76.10                          & 0.811                          & 0.047 \\
\midrule
\rowcolor{gray!15} Ours-B3            & RGB         & \textbf{88.52} & \textbf{0.954}     & \textbf{0.027}   \\ \bottomrule
\end{tabular}%
}
\label{tab:rgbdm}
\vspace{-3mm}
\end{table}


\subsubsection{Comparison on Generic Segmentation}

\vspace{1mm}
\noindent\textbf{Stanford2D3D dataset.}
As shown in Table~\ref{tab:s2d3d_sota_full}, we compare with other methods across different backbone sizes. Our method outperforms existing work by about $10.1\%$ in mIOU, highlighting the segmentation capacity of our network in general scenes with glass objects.


\begin{table}[!t]
\centering
\footnotesize
\caption{\small{Comparison with SOTAs on Stanford2D3D dataset.} 
}
\setlength{\tabcolsep}{4pt}
\renewcommand{\arraystretch}{1.1}
\resizebox{0.46\textwidth}{!}{%
\begin{tabular}{lccc}
\toprule
\textbf{Method} &  \textbf{GFLOPs}~$\downarrow$ & \textbf{MParams}~$\downarrow$ & \textbf{mIoU}~$\uparrow$ \\ 
\midrule \midrule
PVT-T~\cite{pvt2021} & 10.16  & 13.11 & 41.00 \\
Trans4Trans-T~\cite{zhang2022trans4trans} & 10.45  & 12.71 & 41.28 \\
\rowcolor{gray!15} Ours-T &  10.50 & 12.72 & \underline{47.11} \\
Trans2Seg-T~\cite{xie2021segmenting} &  16.96 & 17.87 & 42.07 \\
\rowcolor{gray!15} Ours-B1 &  21.99 & 14.87 & \textbf{51.55} \\
\midrule 
PVT-S~\cite{pvt2021} & 19.58  & 24.36 & 41.89 \\
Trans4Trans-S~\cite{zhang2022trans4trans} & 19.92  & 23.95 & 44.47 \\
\rowcolor{gray!15} Ours-S &  20.00 & 23.98 & \underline{50.17} \\
Trans2Seg-S~\cite{xie2021segmenting} &  30.26 & 27.98 & 42.91 \\
\rowcolor{gray!15} Ours-B2 &  37.03 & 27.59 & \textbf{53.98} \\
\midrule
Trans4Trans-M~\cite{zhang2022trans4trans} & 34.38  & 43.65 & 45.73 \\
\rowcolor{gray!15} Ours-M &  34.51 & 43.70 & \underline{52.57} \\ 
Trans2Seg-M~\cite{xie2021segmenting} &  40.98 & 30.53 & 43.83 \\
PVT-M~\cite{pvt2021} & 49.00  & 56.20 & 42.49 \\
\rowcolor{gray!15} Ours-B3 &  68.35 & 51.21 & \textbf{54.66} \\
\midrule
\rowcolor{gray!15} Ours-L &  50.54 & 60.86 & \underline{53.75} \\
\rowcolor{gray!15} Ours-B4 &  79.34 & 67.11 & \textbf{55.21} \\
\midrule
\rowcolor{gray!15} Ours-B5 &  154.37 & 106.19 & \textbf{55.83} \\
\bottomrule
\end{tabular}
} 
\label{tab:s2d3d_sota_full}
\vspace{-3mm}
\end{table}

\vspace{1mm}
\noindent\textbf{TROSD dataset.}
We compared our method with SOTAs on the TROSD dataset~\cite{trosd2023}, a dataset specifically for transparent and reflective objects. Table~\ref{tab:trosd_full} provides an overview of our competitors and highlights their best results, achieved using their publicly available source codes. All methods utilized the same data augmentation strategy. 

\vspace{1mm}
\noindent\textbf{ADE20k and Cityscapes datasets.}
We conducted additional experiments on the ADE20K and CityScapes datasets, with the results (mIoU) shown in Table~\ref{tab:ade-sc}, sorted in ascending order of GFLOPs ($512 \times 512$). As can be seen, our method performs well on both datasets, with mIoU $47.5\%$ on ADE20K and $81.9\%$ on CityScapes.

\begin{table}[!t]
\centering
\small
\caption{\small{Comparison of different methods on TROSD. R: reflective objects. T: transparent objects. B: background.}}
\setlength{\tabcolsep}{3pt}
\renewcommand{\arraystretch}{1.1}
\resizebox{0.49\textwidth}{!}{%
\begin{tabular}{lcH|ccccc}
\toprule
\multirow{2}{*}{\textbf{Method}} & \multirow{2}{*}{\textbf{Input}} & \multirow{2}{*}{\textbf{Backbone}} & \multicolumn{3}{c}{\textbf{IOU~$\uparrow$}} & \multirow{2}{*}{\textbf{mloU~$\uparrow$}} & \multirow{2}{*}{\textbf{mAcc~$\uparrow$}} \\ 
                                 &                                 &                                    & \textbf{R} & \textbf{T} & \textbf{B} &                                    &                                    \\ 
\midrule \midrule
RefineNet~\cite{refinenet}                        & RGB                             & ResNet-101                         & 21.32      & 37.32      & 92.37      & 50.34                              & 63.59                              \\
ANNNet~\cite{ANNNet}                           & RGB                             & ResNet-101                         & 22.31      & 41.30       & 93.43      & 52.35                              & 62.49                              \\
Trans4Trans~\cite{zhang2022trans4trans}                      & RGB                             & PVTv1                                & 27.69      & 39.22      & 94.16      & 53.69                              & 61.82                              \\
PSPNet~\cite{pspnet}                           & RGB                             & ResNet-101                         & 26.35      & 44.38      & 94.19      & 54.97                              & 64.14                              \\
OCNet~\cite{ocnet}                            & RGB                             & ResNet-101                         & 31.76      & 46.52      & 95.05      & 57.78                              & 64.46                              \\
TransLab~\cite{xie2020segmenting}                         & RGB                             & ResNet-50                          & 42.57      & 50.72      & 96.01      & 63.11                              & 68.72                              \\
DANet~\cite{danet}                            & RGB                             & ResNet-101                         & 42.76      & 54.39      & 95.88      & 64.34                              & 70.95                              \\
TROSNet~\cite{trosd2023}                          & RGB                             & ResNet-50                          & 48.75      & 48.56      & 95.49      & 64.26                              & 75.93                              \\
\rowcolor{gray!15} Ours                             & RGB                             & PVTv2                                & \textbf{66.16}      & \textbf{66.83}      & \textbf{97.71}      & \textbf{76.90}                              & \textbf{87.62}                               \\
\midrule
SSMA~\cite{SSMA}                             & RGB-D                           & ResNet-50                          & 24.70       & 29.04      & 89.98      & 47.91                              & 67.72                              \\
FRNet~\cite{frnet}                            & RGB-D                           & ResNet-34                          & 28.37      & 36.59      & 92.18      & 52.38                              & 63.94                              \\
EMSANet~\cite{emsanet}                          & RGB-D                           & ResNet-101                         & 27.53      & 44.10       & 96.14      & 55.92                              & 71.63                              \\
FuseNet~\cite{fusenet}                          & RGB-D                           & VGG-16                             & 37.30       & 43.29      & 94.97      & 58.52                              & 66.13                              \\
RedNet~\cite{rednet}                           & RGB-D                           & ResNet-50                          & 48.27      & 47.57      & 95.76      & 63.87                              & 69.23                              \\
EBLNet~\cite{He_2021_ICCV}                           & RGB-D                           & ResNet                            & 51.75      & 50.12      & 94.57      & 65.49                              & 67.39                              \\
TROSNet~\cite{trosd2023}                          & RGB-D                           & ResNet-50                          & \underline{62.27}      & \underline{57.23}      & \underline{96.52}      & \underline{72.01}                              & \underline{81.21}                              \\ 
\bottomrule
\end{tabular}%
}
\label{tab:trosd_full}
\end{table}

\begin{table}[!t]
\centering
\footnotesize
\caption{\small{Comparison (mIoU~$\uparrow$) with SOTAs on ADE20k and Cityscapes (CitySc.) datasets.}}
\setlength{\tabcolsep}{3pt}
\renewcommand{\arraystretch}{1.1}
\resizebox{0.49\textwidth}{!}{%
\begin{tabular}{l|ccc|cc}
\toprule
\textbf{Method}                            & \textbf{GFLOPs↓} & \textbf{MParams↓} & \textbf{Backbone} & \textbf{ADE20K}  & \textbf{CitySc.} \\ \midrule
Trans4Trans-M~\cite{wang2021pvtv2}         & 41.9             & 49.6              & PVTv2-B3          & -                & 69.3                \\
Semantic FPN~\cite{zhang2022trans4trans}   & 62.4             & 49.0              & PVTv2-B3          & 47.3             & -                   \\
Ours-B3                                    & 68.3             & 51.2              & PVTv2-B3          & 47.5             & \underline{81.9}    \\
MogaNet-S~\cite{Li2022MogaNet}             & 189              & 29.0              & SemFPN            & \underline{47.7} & -                   \\
NAT-Mini~\cite{Hassani2023NAT}             & 900              & 50.0              & UPerNet           & 46.4             & -                   \\
InternImage-T~\cite{Wang2023Intern}        & 944              & 59.0              & UPerNet           & \textbf{47.9}    & \textbf{82.5}       \\ \bottomrule
\end{tabular}%
}
\label{tab:ade-sc}
\end{table}


\subsection{Ablation studies}
\label{sec:supp_tos_ablation}


We present additional ablation studies to verify various aspects of our model's design.

\vspace{1mm}
\noindent\textbf{Different combinations of network architecture.}
Table~\ref{tab:combination} presents comparisons among various combinations of encoders and decoders, such as using only a CNN architecture, using a combination of CNN and Transformer, and using a fully Transformer-based model. Our method, an encoder-decoder transformer-based model, outperforms competitive networks, indicating the system's capability to segment transparent objects effectively. In this ablation study, we used Ours-M and Ours-B2 (not the best model, Ours-B5), which have the same network size as other methods (-M model size), for a fair comparison.

\begin{table}[!t]
\footnotesize
\centering
\caption{\small{Effectiveness of different network architecture combinations. Models are evaluated on the Trans10K-v2 dataset~\cite{xie2021segmenting}.
\textbf{Note that:} the results are sorted by ascending of GFLOPS. 
}}
\setlength{\tabcolsep}{3pt}
\renewcommand{\arraystretch}{1.1}
\resizebox{0.49\textwidth}{!}{
\begin{tabular}{l|cc|cc|cc}
\toprule
\multirow{2}{*}{\textbf{Method}} & \multicolumn{2}{c|}{\textbf{Encoder}} & \multicolumn{2}{c|}{\textbf{Decoder}} & \multirow{2}{*}{\textbf{GFLOPs}} & \multirow{2}{*}{\textbf{mIoU}} \\
& \textbf{Trans.} & \textbf{CNN} & \textbf{Trans.} & \textbf{CNN} & & \\ 
\midrule \midrule
Trans4Trans-M~\cite{zhang2022trans4trans} & \checkmark & & \checkmark & & 34.3 & 75.1 \\
\rowcolor{gray!15} Ours-M                                    & \checkmark & & \checkmark & & 34.5 & \underline{76.1} \\
\rowcolor{gray!15} Ours-B2                                   & \checkmark & & \checkmark & & 37.0 & \textbf{79.3} \\
Trans2Seg-M~\cite{xie2021segmenting}      & & \checkmark & \checkmark & & 40.9 & 69.2 \\ 
FCN~\cite{fcn2015}                            & & \checkmark & & \checkmark & 42.2 & 62.7 \\
OCNet~\cite{ocnet}                        & & \checkmark & & \checkmark & 43.3 & 66.3 \\
PVT-M~\cite{pvt2021}                      & \checkmark & & \checkmark & & 49.0 & 72.1 \\
\bottomrule
\end{tabular}
}
\label{tab:combination}
\end{table}

\begin{figure}[!t]
    \centering
    \includegraphics[width=0.86\linewidth]{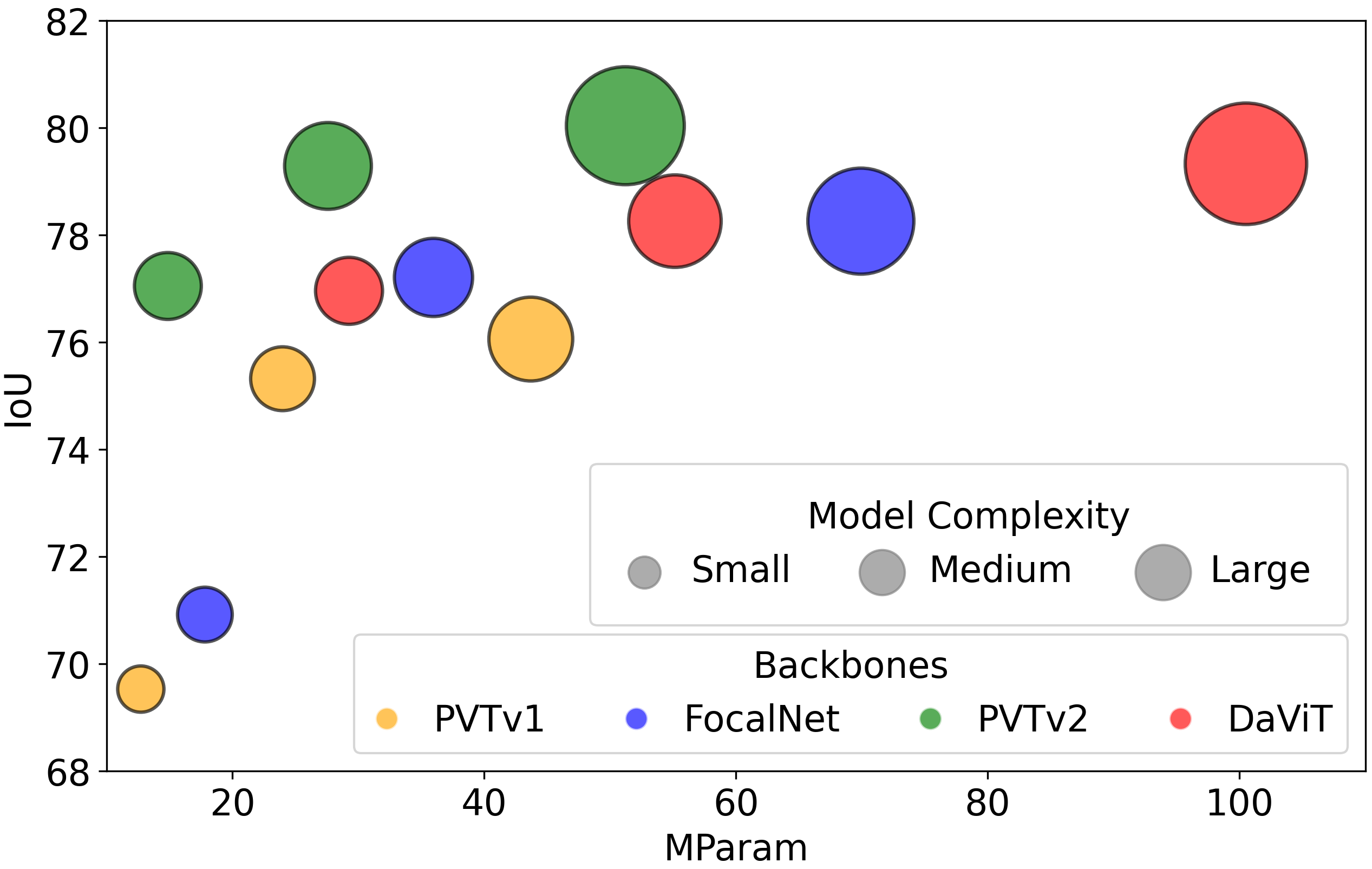}
    \caption{\small{Our method with various backbones on Trans10K-v2 dataset. The bubble's size is its complexity in GFLOPs.}}
    \label{fig:backbones}
    \vspace{-2mm}
\end{figure}

\vspace{1mm}
\noindent\textbf{Analysis of different backbones.} 
We have conducted experiments using alternative backbones, as presented in Figure~\ref{fig:backbones}. Among these options, the PVT-v2 backbone~\cite{wang2021pvtv2} stands out with significantly higher mIoU and remarkably compact model size (MParams). Despite its higher GFLOP complexity compared to the FocalNet backbone~\cite{yang2022focal}, it still achieves better performance. Additionally, the PVT-v2 backbone~\cite{wang2021pvtv2} demonstrates a lower complexity than the DaViT backbone~\cite{ding2022davit} while maintaining competitive mIoU results. These findings highlight the superiority of the PVT-v2 backbone~\cite{wang2021pvtv2} in achieving an optimal balance between performance and model size, making it a promising choice for our method. When comparing PVT-v1~\cite{pvt2021} with other backbones, the PVT-v1~\cite{pvt2021} backbone boasts a considerably smaller model size and lower complexity. Despite these advantages, its performance remains competitive with other backbones. This demonstrates the efficiency of the PVT-v1 backbone~\cite{pvt2021}, which delivers comparable performance while being lighter and less computationally demanding. 

\vspace{-2mm}
\subsection{Further Analysis and Discussions}
\label{sec:further_analysis}

\begin{figure}[!t]
    \centering
    \includegraphics[width=0.94\linewidth]{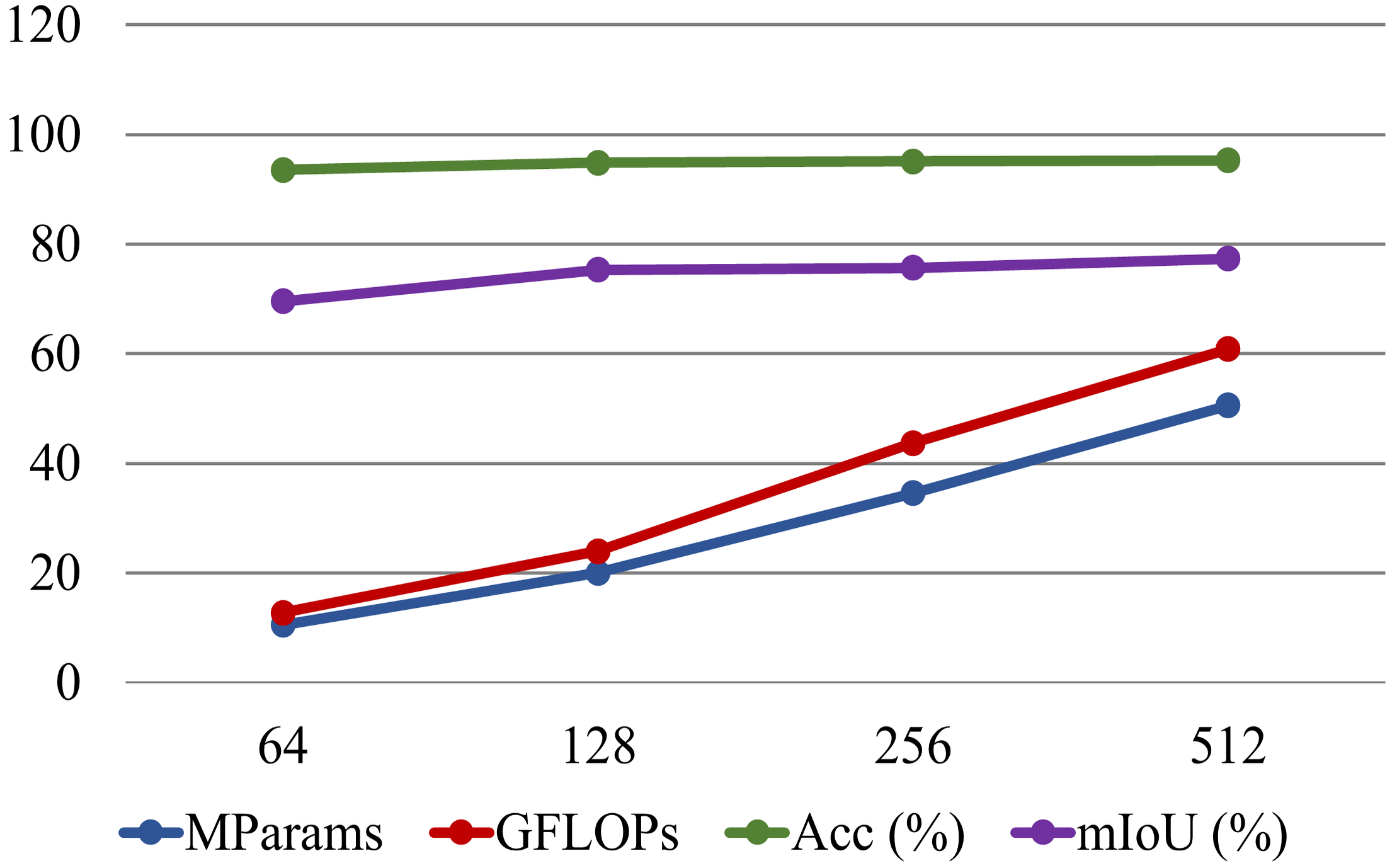}
    \caption{\small{Effectiveness of the embedding channel in our method on Trans10K-v2~\cite{xie2021segmenting}.}}
    \label{fig:channel_test_miou}
    \vspace{-3mm}
\end{figure}

\paragraph{Effectiveness of the embedding channel.} We experiment with the embedding channel with various values (64, 128, 256, 512) and report the mIoU and Accuracy of Ours-B1 model in Figure~\ref{fig:channel_test_miou}. Throughout the results, we show that our model achieved better performance with a higher number of embedding channels (from 77.05\% at 64 channels to 78.85\% at 512 channels). Note that, due to memory limits, we cannot perform experiments with higher embedding channels, \eg, 1024 or 2048, and to save computational resources, we used Ours-B1 in this ablation study.

\begin{figure}[!t]
    \centering
    \includegraphics[width=0.94\linewidth]{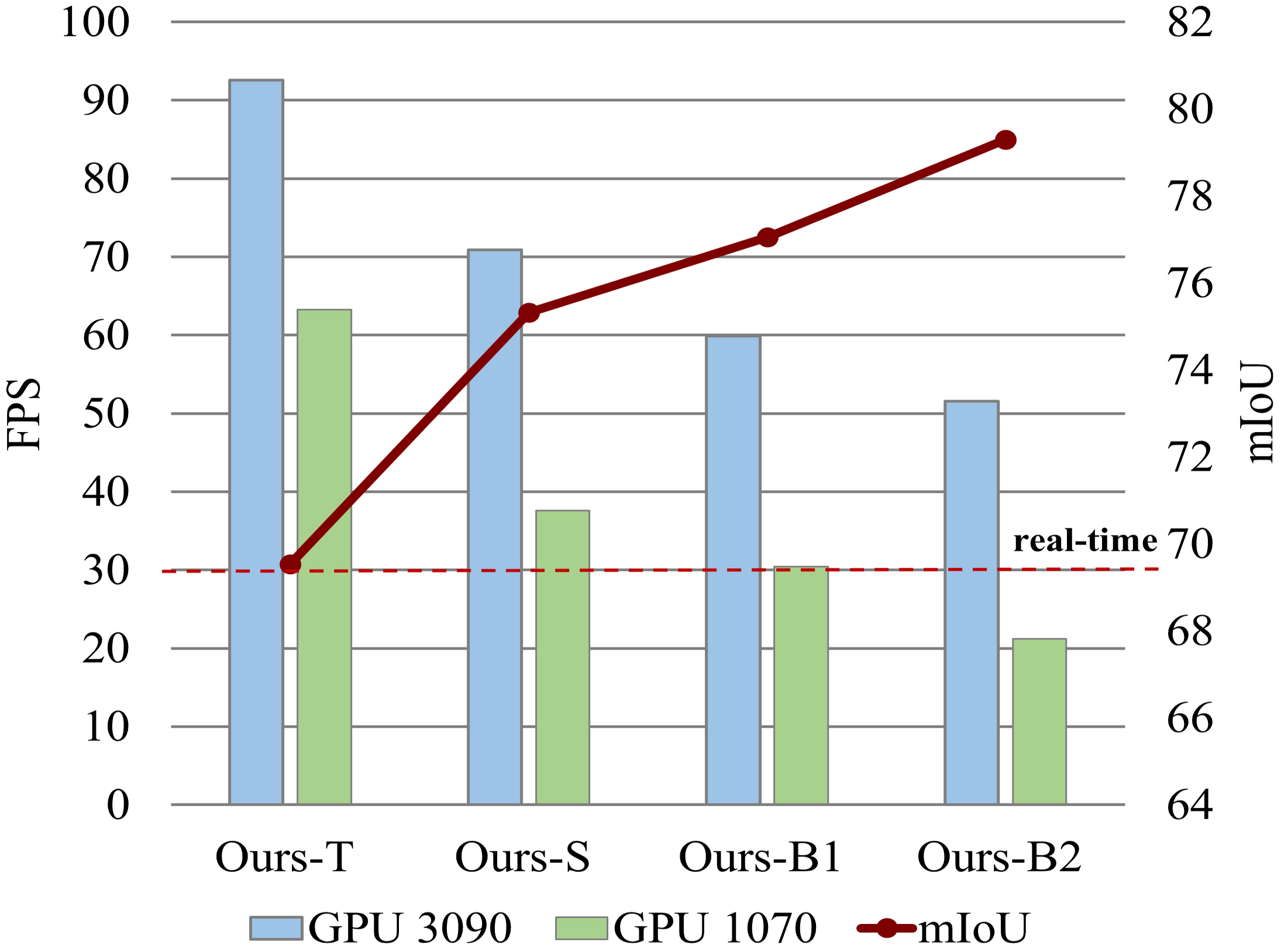}
    \caption{\small{Inference time (FPS) of our method on Trans10K-v2.}}
    \label{fig:time}
    \vspace{-3mm}
\end{figure}

\paragraph{Real-time performance.} We report the inference speed of our models on different GPUs (NVIDIA GTX 1070, NVIDIA RTX 3090) at a resolution of $512 \times 512$ and a batch size of 1. As shown in Figure~\ref{fig:time}, while Our-T model has a lower computational cost than other versions, it's important to note that all these models deliver performance levels well-suited for deployment on robotic systems. In real-world situations, achieving a similar level of prediction accuracy for each frame is crucial because it enables a navigation system to be more responsive, improving its capacity to assist robots efficiently.

\paragraph{Incorporation with other modalities.} Integrating our model with depth images (RGB-D) or polarization images (RGB-P) is a feasible enhancement. A naive method involves adding an extra encoder to extract features from depth or polarization data. These additional features would then be fused with RGB features before our FPM module. This strategy is in line with PDNet~\cite{rgbdm2021}, TROSNet~\cite{trosd2023}, and PGSNet~\cite{PGSNet2022}, as detailed in the supplementary material. Notably, including depth or polarization data in these models has led to significant performance gains and increased computational costs. Specifically, with added depth information, PDNet and TROSNet improved by $+4\%$ and $+8\%$ mIoU, and with added polarization information, PGSNet experienced a $+5\%$ boost in mIoU.

\paragraph{Utilizing reflection removal methods for detecting reflections.} Employing reflection removal techniques, as discussed in recent studies~\cite{Dong2021Removal,Song2023removal}, offers the potential to generate pseudo labels with distinct advantages. However, these methods are mainly designed to address global reflections when an image is entirely encompassed by glass. These methods have limitations in complex real-world situations in which glass objects are distributed throughout the scene rather than occupying a dominant position. Our study introduces the RFE module, which can detect localized reflections and distinguish glass surfaces based on the semantic mask. This module is better suited to the diverse and unpredictable conditions found in real-world situations, where reflections are specific to certain areas rather than uniformly distributed across the entire image, making it a better fit for real-world scenarios.

\begin{figure}[!t]
    \centering
    \includegraphics[width=0.98\linewidth]{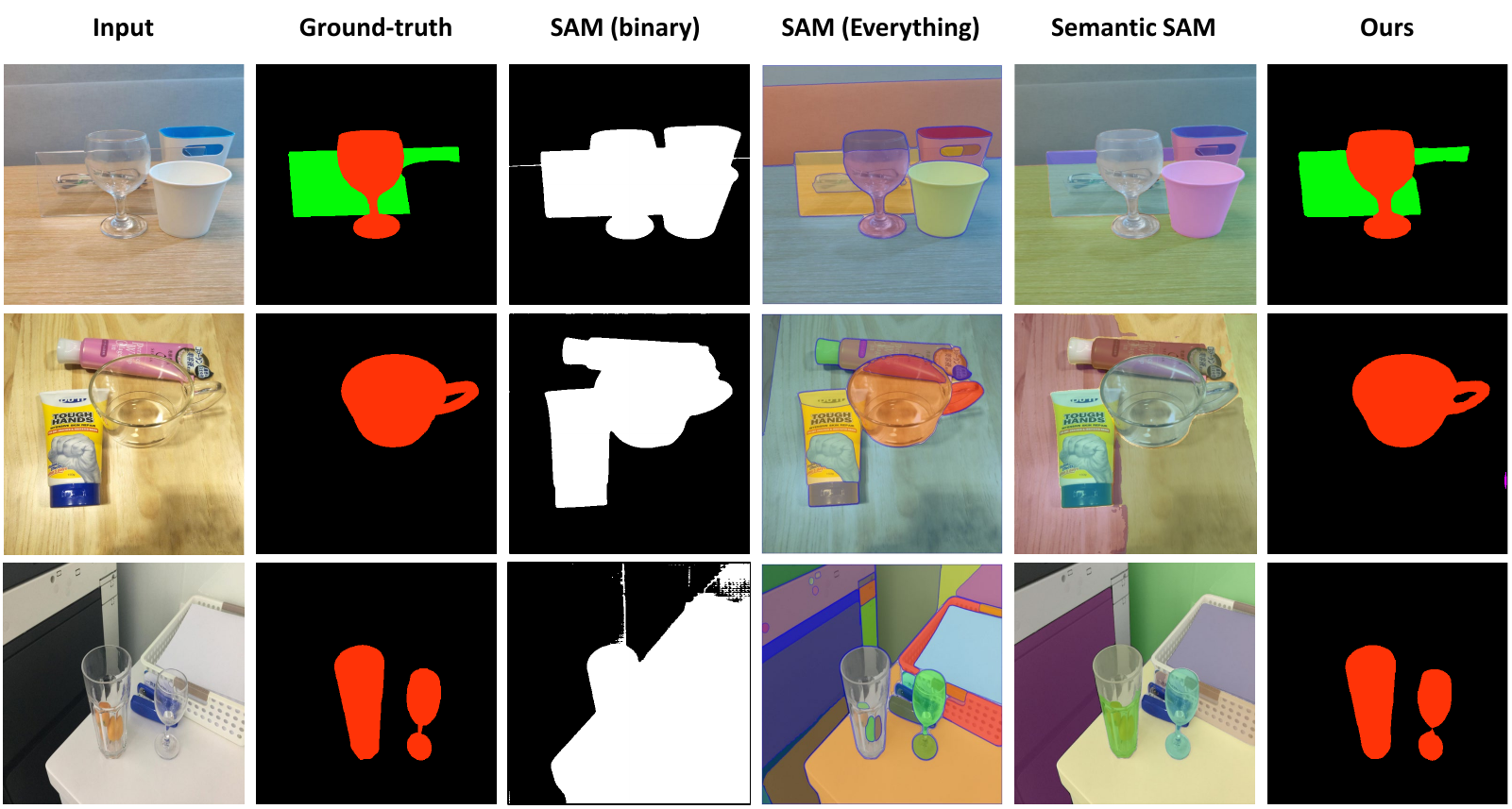}
    \caption{\small{Qualitative comparison of our method with recent foundation models on Trans10k-v2 dataset.}}
    \label{fig:sam}
    \vspace{-3mm}
\end{figure}

\paragraph{Comparison with foundation models.}
To fully evaluate our method's performance, we also compared it with recent powerful foundation models, such as the SAM model, and the results are shown in Figure~\ref{fig:sam}. It is important to note that the SAM model \textbf{does not include semantics, or in other words, it cannot yield masks with semantics}. The SAM model also presents the challenge of over-segmentation, thereby increasing the likelihood of false positives. As a result, we see that the SAM model (binary and everything) cannot distinguish between glass and non-glass regions, unlike our method. It is the same for SAM variants with semantics~\cite{li2023semanticsam,chen2023semantic}, which still fail and cannot generate reasonable semantics either.


\begin{figure}[!t]
    \centering
    \includegraphics[width=0.94\linewidth]{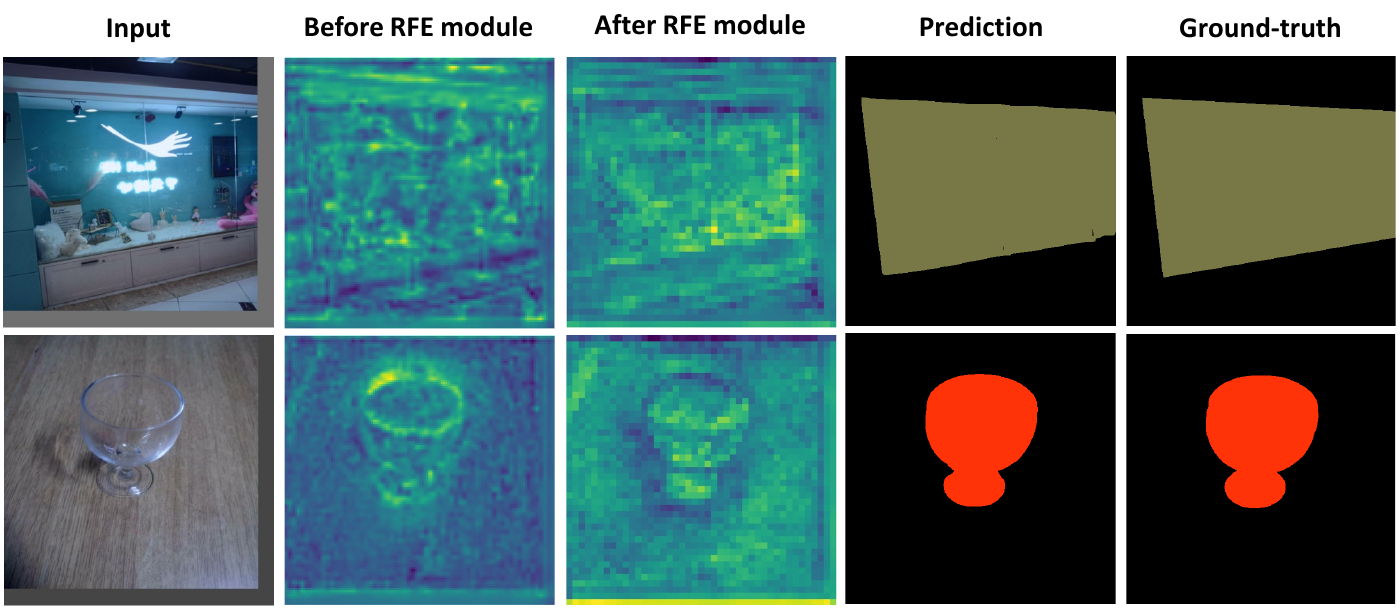}
    \caption{\small{Comparison of the feature maps before and after passing through the RFE module on the Trans10k-v2 dataset.}}
    \label{fig:rfe}
    \vspace{-3mm}
\end{figure}

\begin{figure}[!t]
    \centering
    \includegraphics[width=0.94\linewidth]{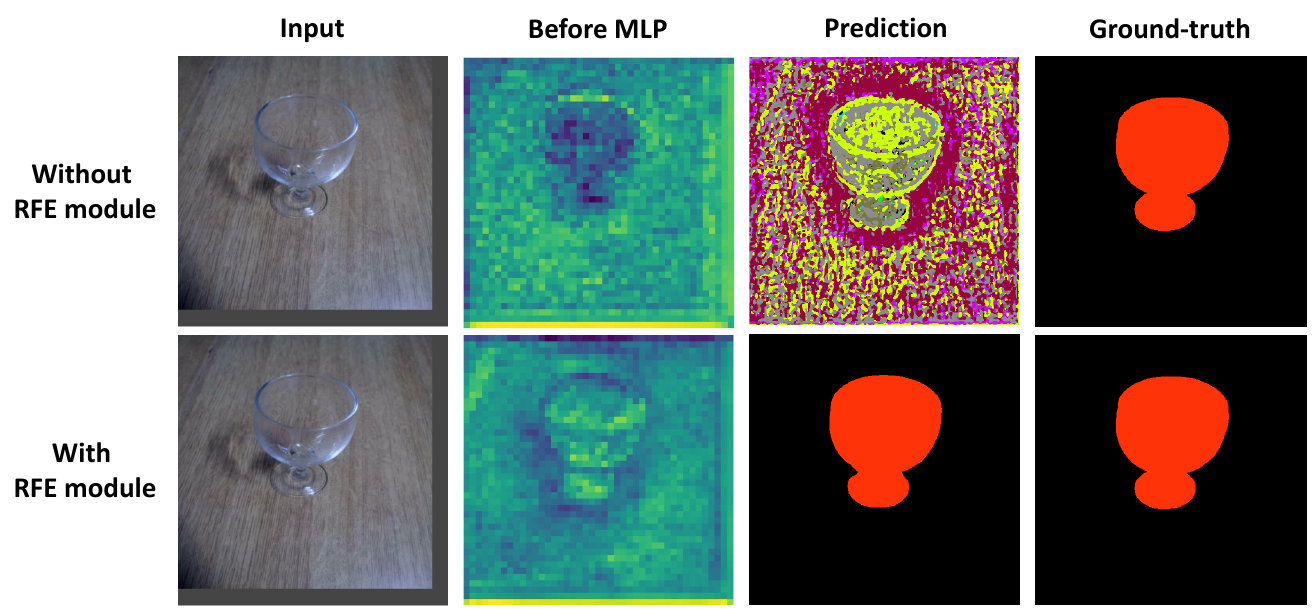}
    \caption{\small{Comparison of the feature maps at inference by passing through the RFE module as usual (top) and bypassing the RFE module (bottom) on the Trans10k-v2 dataset.}}
    \label{fig:no_rfe}
    \vspace{-3mm}
\end{figure}

\paragraph{Further analysis on our reflection RFE module.}
To provide further verification of the effectiveness of the RFE module, we conducted the following additional experiments:
\begin{itemize}
    \item We take a model with the RFE module that has already been trained. To demonstrate the effectiveness of RFE, we compare the feature maps before and after passing through the RFE module. In Figure~\ref{fig:rfe}, we can see that after passing through the RFE module, we can get a stronger reflection signal, such as the transparent glass area or the specular reflection appearing at the base of the wine glass.
    \item Using the same model, we try turning off the RFE module at inference by passing the feature map before RFE directly to the next step. Note that at training, the RFE module is \textit{well-trained as usual}. Figure~\ref{fig:no_rfe} shows that bypassing RFE results in a noisy feature map and wrong mask prediction. This means that our learning of RFE does not yield a trivial function, \eg the identity, and that RFE plays an important role in processing the feature maps and output reflection masks.  
\end{itemize}

{
    \small
    \bibliographystyle{ieeenat_fullname}
    \bibliography{main}
}

\end{document}